\documentclass[11pt]{article}

\usepackage[preprint]{acl}

\usepackage{times}
\usepackage{latexsym}

\usepackage[T1]{fontenc}

\usepackage[utf8]{inputenc}

\usepackage{microtype}

\usepackage{inconsolata}

\usepackage{graphicx}

\usepackage{hyperref}       
\usepackage{url}            
\usepackage{booktabs}       
\usepackage{amsfonts}       
\usepackage{nicefrac}       
\usepackage{microtype}      
\usepackage{xcolor}         

\usepackage[utf8]{inputenc} 
\usepackage[T1]{fontenc}    
\usepackage{hyperref}       
\usepackage{url}            
\usepackage{booktabs}       
\usepackage{amsfonts}       
\usepackage{nicefrac}       
\usepackage{microtype}      
\usepackage{xcolor}         

\usepackage{amssymb} 
\usepackage{amsmath} 
\usepackage{makecell} 
\usepackage{wrapfig} 

\usepackage{graphicx}
\usepackage{pdfpages}
\usepackage{url}
\usepackage{xcolor}
\usepackage{epsfig}
\usepackage{adjustbox}
\usepackage{amsfonts}
\usepackage{amsmath}
\usepackage{amssymb}
\usepackage{booktabs} 
\usepackage{comment}
\usepackage{multirow}
\usepackage{caption}
\usepackage{subcaption}
\usepackage{textcomp}
\usepackage{relsize}
\usepackage{stmaryrd}
\usepackage{rotating}
\usepackage{helvet}
\usepackage{courier}
\usepackage{natbib}
\usepackage{lipsum}
\usepackage{tcolorbox} 
\usepackage{hyperref}
\usepackage{bbm}
\usepackage{algorithm}
\usepackage{algpseudocode}
\usepackage{colortbl} 
\usepackage{arydshln} 
\usepackage{tablefootnote}
\usepackage{mathrsfs}
\usepackage{cleveref}
\makeatletter
\AtBeginDocument{%
  \renewcommand{\sectionautorefname}{\S\@gobble}

  \renewcommand{\subsectionautorefname}{\S\@gobble}  
  \renewcommand{\subsubsectionautorefname}{\S\@gobble}  
    \renewcommand{\appendixautorefname}{\S\@gobble}
}
\makeatother

\definecolor{darkblue}{rgb}{0,0.08,0.5}
\hypersetup{citecolor=darkblue, linkcolor=darkblue, colorlinks = true}

\definecolor{Gray}{gray}{0.85}

\title{Quantifying Genuine Awareness in Hallucination Prediction\\
Beyond Question-Side Shortcuts}

\author{
Yeongbin Seo~~~~~~~~~~~~~~~~~~
Dongha Lee ~~~~~~~~~~~~~~~~~~
Jinyoung Yeo  \thanks{Corresponding author}\\
Department of Artificial Intelligence \\
Yonsei University\\
\texttt{\{suhcrates,donalee,jinyeo\}@yonsei.ac.kr}\\    
}

\begin{document}

\maketitle

\begin{abstract}
Many works have proposed methodologies for language model (LM) hallucination detection and reported seemingly strong performance. However, we argue that the reported performance to date reflects not only a model’s genuine awareness of its internal information, but also awareness derived purely from question-side information (e.g., benchmark hacking). While benchmark hacking can be effective for boosting hallucination detection score on existing benchmarks, it does not generalize to out-of-domain settings and practical usage. Nevertheless, disentangling how much of a model’s hallucination detection performance arises from question-side awareness is non-trivial. To address this, we propose a methodology for measuring this effect without requiring human labor, Approximate Question-side Effect (AQE). Our analysis using AQE reveals that existing hallucination detection methods rely heavily on benchmark hacking.
 The code is available online (\url{https://github.com/ybseo-ac/AQE}).
\end{abstract}

\section{Introduction}

The defining and quantifying of human-like mental attributes in large language models (LLMs) lie at the heart of a long-standing question: whether artificial systems can possess minds akin to our own. While recent advances show that LLMs can rival or even surpass humans in rational reasoning tasks \citep{brown2020languagemodelsfewshotlearners,grattafiori2024llama, openai2023gpt}, higher-order traits such as self-awareness and emotion remain poorly understood, partly due to ambiguities in their definition and measurement \citep{li2024quantifying, yin2023largelanguagemodelsknow}.

Among these traits, self-awareness of knowledge is particularly important because of its close connection to hallucination detection, which is critical for the reliability of LLMs. Although hallucination can arise from various sources, a major cause is answering questions beyond the model’s pre-trained knowledge \citep{tonmoy2024comprehensive}. Humans can recognize when they lack relevant knowledge and refrain from answering \citep{irak2019neurobiological, koriat1993we}, whereas LLMs lack such awareness and tend to generate plausible outputs regardless, leading to hallucination.

Then, how can we define and measure self-awareness of LLMs?
Prior work has often equated self-awareness with hallucination detection itself, motivated by its practical importance. Indeed, recent studies report high hallucination detection performance \citep{snyder2024early, zhang2024rtuninginstructinglargelanguage, manakul2023selfcheckgptzeroresourceblackboxhallucination, azaria2023internal}. 

However, we argue that hallucination prediction does not directly measure self-awareness, because two distinct sources of information are typically involved in the prediction process: (1) information about the model itself and (2) information about the question. As such, hallucination prediction reflects a mixture of \textbf{model-awareness (self-awareness)} and \textbf{question-awareness}. To isolate self-awareness, we disentangle these two components and introduce a Shapley-based metric, the Approximate Question-side Effect (AQE), to quantify question-awareness. The contribution of self-awareness is then estimated by subtracting AQE from hallucination detection precision.

Quantifying self-awareness has important practical implications. As shown in \autoref{sec:aqe}, hallucination predictors that rely heavily on question-awareness often exploit dataset-specific shortcuts and fail to generalize under distribution shifts. In contrast, approaches grounded in model-side information yield more robust performance. We empirically support this claim through dataset analyses and experiments in \autoref{app:case_study}, \autoref{sec:exp_aqe}, and \autoref{sec:experiment}.


Lastly, we also propose a method to enhance the use of model-side information, by leveraging the confidence scores of LLMs more effectively. The proposed method is \textbf{S}emantic \textbf{C}ompression by \textbf{A}nswering in \textbf{O}ne word (SCAO). We demonstrate that SCAO performs particularly well in low AQE settings. Though this method shows limitations in certain settings, such as long-form question answering, it provides clues to overcoming the limitations of previous probing-based methods.

Our contributions are summarized as follows:
\begin{itemize}
\item Conceptual: We disentangle hallucination detection into self-awareness and question-awareness and provide a measurable definition of self-awareness in LLMs.
\item Methodological: We introduce AQE, a Shapley-based metric to quantify question-side effects.
\item Empirical: We show that shortcut-driven, question-aware methods fail to generalize, while model-side approaches are more robust.
\end{itemize}

\section{Related works}
\label{app:related}
As human self-awareness has been extensively studied in cognitive psychology and neuroscience, its core mechanisms can be leveraged to structure and categorize approaches for evaluating internal confidence and hallucination in LLMs.

\paragraph{Self-awareness in humans: Insights from cognitive neuropsychology}

Extensive research in cognitive psychology and neuroscience has shown that human self-awareness—particularly in the context of knowing whether one knows something—relies on layered cognitive processes. According to studies such as \cite{koriat1993we, irak2019neurobiological, brown2017confabulation}, two key mechanisms underpin this self-assessment. 

\textbf{1) Unconscious level:} When a query is received, the brain initiates implicit memory retrieval, evaluating whether candidate memories fit the contextual cues. This process activates regions such as the orbitofrontal and prefrontal cortices within 300–500 milliseconds \citep{schnider2001spontaneous, irak2019neurobiological}, distinct from areas responsible for linguistic output, such as the posterior temporal lobe (i.e., Wernicke’s area) \citep{binder2015wernicke}. Several factors—such as the amount, accessibility, and vividness of retrieved information—act as internal cues signaling potential knowledge \citep{koriat1993we}.

\textbf{2) Conscious level:} The results of these unconscious processes are then consciously evaluated through metacognitive strategies. These include checking for logical and temporal consistency or aligning retrieved information with known frameworks. This layered evaluation reflects human self-awareness—our capacity to introspect on our own knowledge states and confidence levels.

The dual-process theory \citep{kahneman2011thinking} offers a broader framing: rapid, intuitive processes dominate simple recall tasks, while complex, deliberative processes support tasks such as reasoning or problem-solving.

\paragraph{Self-awareness in LLMs: A perspective on hallucination detection}

In large language models (LLMs), the concept of self-awareness can be reinterpreted as the model’s ability to internally assess whether it possesses sufficient knowledge to answer a question accurately. This introspective capacity—whether performed before or after answer generation—is directly related to hallucination detection mechanisms. Our analysis proposes aligning hallucination mitigation strategies with the structure of human self-awareness processes.

\textbf{1) Before-generation:} Approaches that attempt to detect potential hallucinations *before* the model generates an answer mirror the unconscious processes of humans. These methods, including our own, rely on internal indicators such as activation patterns or uncertainty proxies to determine knowledge sufficiency prior to verbalization \citep{mallen2022not}. Benchmarks like Mintaka \citep{sen2022mintakacomplexnaturalmultilingual} and ParaRel \citep{elazar2021pararel} primarily test immediate factual recall, aligning well with this layer of self-assessment.

\textbf{2) After-generation:} Other approaches evaluate the model’s response \textbf{after} generation, resembling conscious-level reasoning in humans. These include multi-pass generation, self-consistency checks, and the integration of external tools such as retrievers \citep{béchard2024reducinghallucinationstructuredoutputs, manakul2023selfcheckgptzeroresourceblackboxhallucination, chen2024insidellmsinternalstates}. Benchmarks requiring structured reasoning, such as MMLU \citep{hendrycks2021measuringmassivemultitasklanguage}, TruthfulQA \citep{lin2022truthfulqameasuringmodelsmimic}, and ELI5 \citep{fan2019eli5}, are more aligned with this stage, as they demand deliberative thought processes rather than pure retrieval.

Importantly, not all hallucinations can be attributed to failures in self-awareness. For instance, hallucinations in open-book QA tasks may arise from comprehension or inference errors rather than epistemic uncertainty. Benchmarks like SQuAD \citep{rajpurkar2016squad100000questionsmachine} and FEVER \citep{thorne2018fever} exemplify this issue, being more about information grounding than self-assessment.

Overall, hallucination in LLMs encapsulates a variety of cognitive failures. Addressing them requires different forms of internal awareness—ranging from assessing memory sufficiency to verifying logical coherence. Ultimately, a robust system may need to combine multiple self-assessment mechanisms. In this work, we focus on introspective strategies relevant to knowledge recall before answer generation, drawing inspiration from unconscious-level self-awareness in humans.

\section{Definition}

\label{sec:definition}
In this section, we first examine the definition of self-awareness of human. Next, we define the task formulation of hallucination prediction, to establish a definition of self-awareness in LLMs. And we review the definitions from previous works. More detailed review on the related works of neuropsychology and LLMs is in \autoref{app:related}.

\paragraph{Self-awareness of human \ \  }
In psychology, human self-awareness is defined as the capability of perception of one's own mental processes or states, which includes thoughts, feelings, emotions, and knowing \citep{morin2011self}. In this work, we focus on the self-awareness of knowing certain knowledge, which is also referred to as self-knowledge \citep{yin2023largelanguagemodelsknow}.

Some studies on LLMs also borrow the term ``meta-cognition'' from psychology to refer to self-knowledge \citep{liteach}, which may be inaccurate. It is because meta-cognition focuses on the conscious level \citep{koriat1993we}, while recent research suggests that the human brain conducts judgment of knowing even at the unconscious level \citep{irak2019neurobiological}, which is even before consciously recognizing the meaning of the question.

\paragraph{Hallucination prediction }
The term ``hallucination'' has been broadly used to refer to the phenomenon where a model provides an incorrect answer to a given question \citep{li2023haluevallargescalehallucinationevaluation, manakul2023selfcheckgptzeroresourceblackboxhallucination, béchard2024reducinghallucinationstructuredoutputs}. Thus, ``hallucination detection'' refers to the task of predicting whether a response is incorrect \citep{li2023haluevallargescalehallucinationevaluation, chen2024insidellmsinternalstates}. 

In this work, we focus specifically on (1) hallucination from \textbf{factoid questions} that examine whether the model possesses certain knowledge, as this has been widely used as a clear and straightforward scenario for exploring hallucination detection \citep{snyder2024early, zhang2024rtuninginstructinglargelanguage}. (2) And we focus on an early detection (i.e., \textbf{prediction}) scenario \citep{snyder2024early, chen2024insidellmsinternalstates, azaria2023internal} where \( k \) is predicted before answer generation, as this setting is more appropriate for examine self-awareness (we describe this in the next section).

To formulate the common problem setting of hallucination prediction, let $\theta$ represent the model, $x$ the query, and $y$ the answer label. The model $\theta$ infers $\hat{y}$ from the input $x$. The correctness of $\hat{y}$ can then be measured by evaluating its similarity to $y$, denoted as $k$, representing a binary value (True/False). Common evaluation methods include string matching \citep{zhang2024rtuninginstructinglargelanguage}, GLUE, and G-eval \citep{liu2023gevalnlgevaluationusing}. During this process, a datapoint \( s_{\theta, x}=\theta(x) \) is extracted from $\theta$, which contains information about how \( \theta \) perceived \( x \). We denote this as $s$ for simplicity. Through a series of question-answering and evaluation processes, we obtain a dataset $\mathcal{D} = \{(s_i, k_i) \}_{i=1}^{N}$, where $N$ is the dataset size. From this, hallucination prediction is defined as a binary classification task where a learnable module $\phi$ learn to take input $s_i$ to predict $k_i$, which we note $\hat{k_i}=\phi(s_i)$. 

\begin{figure*}[t!]
\begin{center}
  \includegraphics[width=0.8\linewidth]{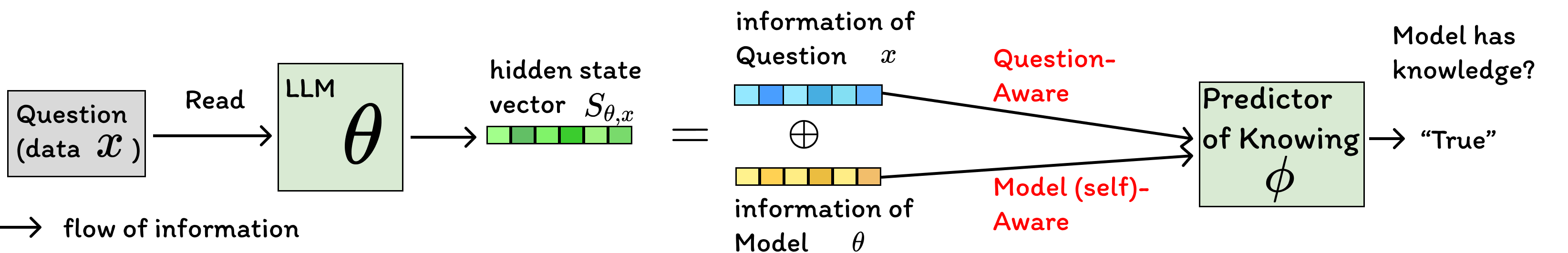}
  \end{center}
  \caption{Pipeline for the prediction of knowing (prediction of hallucination).  }
  \label{fig:pipeline}
\vspace{-10pt}
\end{figure*}


Assuming  $\theta$ as a transformer model \citep{vaswani2023attentionneed}, $ s$ can be mainly two forms (1) Hidden state vectors: Transformer models are composed of multiple attention layers, where each layer passes a fixed-size vector (i.e., hidden states) to the next. These vectors encode the semantic and contextual information of the input \( x \) \citep{reimers2019sentencebertsentenceembeddingsusing}, and are also known to contain information about the model's response that will be generated \citep{li2024inference}, providing a cue for hallucination prediction.
(2) Confidence score: This refers to the softmax probability that the causal LLMs predict for the next token generation. While the hidden state is a high-dimensional representation (4096 dimensions in \texttt{LLaMA-3-8B}), the confidence score is a scalar value. As the hidden state contains richer information, it has achieved the best performance as a source for hallucination prediction and has been regarded as the main source \citep{snyder2024early, chen2024insidellmsinternalstates}.
\subsection{Formulating the self-awareness of LLM}
\label{sec:self-aware-definition}

We defined hallucination prediction as the process where $\phi$ perceives $s$ to infer $\hat{k}$ to predict $k$. 
As \( s \) represents the state of the model after it has seen a question, it inherently contains information of two distinct objects, the question-side and the model-side, as illustrated in \autoref{fig:pipeline}. The question-side refers to the objective information that can be shared across different models, such as the domain of question (e.g., science, math, society) and the type of question (e.g., multiple choice, open-ended). For human, this type of information is derived from rational abilities (e.g., classification and reading comprehension), rather than from higher-order mental capacity such as self-awareness \citep{morin2011self}. In contrast, the model-side information refers to model-specific attributes, such as the possession of the knowledge that is needed for response, or the degree of confidence for answering. In humans, this corresponds to the domain of self-awareness.

Let us denote the representation of the question-side information in $s$ as $s_Q$, and the representation of the model-side information as $s_M$. We rewrite previously defined hallucination prediction $\hat{k} = \phi(s)$ as $\hat{k} = \phi(s_Q, s_M)$. When we denote the information contained in $s$ as $s$ itself, we can also note $s = s_Q \cup s_M$. This decomposition forms the basis for applying the Shapley value formulation.

Prior studies have also empirically shown that the hidden states of transformer models encode multiple properties in a linearly separable manner. For instance, \citet{li2024inference} demonstrates that the hidden state contains a direction associated with the attribute of ``truthfulness'', therefore linearly adding this to the hidden states results in more truthful responses.

When \( \phi \) learns to predict \( k \) using two sources of information (\( s_Q, s_M \)), utilizing question-side information can be regarded as question-awareness, while utilizing model-side information can be regarded as model-awareness, which is ``self-awareness'' in the perspective of model.
Thus, self-awareness can be formally expressed as: 
\vspace{-5pt}
\begin{eqnarray}
\label{eq:selfaware}
\hat{k} = \phi(s_M).
\end{eqnarray}
\vspace{-20pt}



\textbf{Why hallucination prediction, not detection? \  }
We argue that hallucination prediction is a more suitable setting than hallucination detection for examining self-awareness. In hallucination detection, $\phi$ perceives model-generated answers when predicting \( \hat{k} \), which can be formulated as \( \hat{k} = \phi(s_M, s_Q, x, \hat{y}) \), where \( x \) is the question and \( \hat{y} \) is the generated answer. As self-awareness is defined as \( \hat{k} = \phi(s_M) \), additional inputs \( x \) and \( y \) serve as distracting factors, making it difficult to isolate the effect of \( s_M \). Intuitively, this detection scenario may become more of a reading comprehension task over \( x \) and \( y \), rather than assessing the model's internal states. Therefore, we choose 
hallucination prediction scenario to more clearly examine the effect of self-awareness.


\subsection{Definition from previous works}
\textbf{Utilizing $s$ as self-awareness \ }
Previous works implicitly regard self-awareness as the hallucination detection itself. In other words, the focus has been on the act of predicting $k$ from $s$, with no consideration given to the decomposition of $s$ into $s_Q$ and $s_M$. As a result, some of the hallucination detection performance reported in those works is partially overestimated by the effect of question-side shortcuts  \citep{zhang2024rtuninginstructinglargelanguage, azaria2023internal}. We analyze such cases in \autoref{sec:case_study}.

\textbf{Utilizing $s_Q$ as self-awareness \  }
In another case, some works define utilization of $s_Q$ as self-awareness, which is the opposite concept from our view. \cite{yin2023largelanguagemodelsknow} defines the term ``self-knowledge'' as a self-awareness on possession of certain knowledge. And from this definition, they construct the dataset SelfAware to measure the self-awareness capacity. While the definition in the paper is consistent with ours, the construction of the dataset stands on the opposite definition.

The dataset SelfAware consists of ``answerable'' and ``unanswerable'' questions, and the capacity of self-awareness is defined as the classification between the two. An unanswerable question refers to one that is philosophical (e.g., "What is a happy life?") or subjective (“Do you like to go to the mountains?”), where no definitive answer can be given, thus inevitably leading to hallucination. This setting is contradicting the term ``self-awareness'' in three points: (1) the unanswerability defined in the SelfAware involves no model-side information. For example, the question “Do you like to go to the mountains?” is always classified as ``unanswerable'' in SelfAware, regardless of how much knowledge about mountains is stored in the answerer (e.g., 1B LM, 70B LM, or a human). 

(2) The dataset includes fixed labels indicating the unanswerability of each question, therefore unanswerability is entirely independent from the model's state. As unanswerability represents whether the model can answer a given question, it aligns with $k$ in our problem setting (\autoref{eq:selfaware}). However, as $k$ in SelfAware is determined entirely by properties of the question and not the model's state, the task proposed in this dataset becomes \( k = \phi(s_Q) \), excluding $s_M$. This problem setting is a question-awareness, not a self-awareness, from our definition. (3) For humans, the ability needed for this classification task is reading comprehension, which is not self-awareness.

\section{AQE: assessing question-side effects of hallucination prediction datasets}

As question-awareness is using and relying on the question-side information, in this section, we first identify the data-specific shortcuts that cause dependency on the question-side information, through a case study in existing datasets for hallucination prediction. Next, we introduce our novel metric AQE, a method for quantifying the effect of question-side shortcuts in that dataset.%

\subsection{Case study on question-side shortcuts}
\label{sec:case_study}

We investigate sources of question-side shortcuts in datasets used in hallucination prediction studies. We focus on short-form closed-book factoid question answering scenarios, which include ParaRel \citep{elazar2021pararel}, Mintaka \citep{sen2022mintakacomplexnaturalmultilingual}, HaluEval \citep{li2023haluevallargescalehallucinationevaluation}, HotpotQA \citep{yang2018hotpotqadatasetdiverseexplainable}, and SimpleQuestion \citep{yin2016simplequestionansweringattentive}. We describe three main sources of question-side shortcuts, and further in \autoref{app:case_study}.

\textbf{(1) Broken Question Problem \ }
Many datasets have incomplete annotations for one-to-many question–answer pairs (e.g., Daniel Bernoulli → only “physics” labeled).
This causes correct answers outside the label set to be marked as hallucinations, making the predictor $\phi$ learn domain-based shortcuts rather than true self-awareness.
Datasets like Mintaka fix this by adding constraints for one-to-one mappings.

\textbf{(2) Domain Shortcut \ }
The likelihood of hallucination (k=True) often varies by domain.
For instance, if a model is strong in science but weak in history, history questions will naturally show higher hallucination rates.
Thus, $\phi$ may simply learn domain-based correlations — predicting (k) based on what the question is about rather than what the model actually knows.
This makes the task “question-aware” instead of “model-aware,” undermining the goal of self-awareness.
A truly self-aware model should, like a human, recognize when it specifically knows or doesn’t know something, even within an unfamiliar domain.

\textbf{(3) Question Type Shortcut \ }
The structure of the question itself (e.g., binary-choice, multiple-choice, open-ended) also influences (k).
Binary-choice questions have a higher baseline chance of being correct $p(k=True)\geq 0.5$ even with random guessing, unlike open-ended ones $p(k=True)=0$.
Consequently, $\phi$ might exploit this and always predict “True” for binary-choice questions, forming another non-self-aware shortcut.
Datasets such as HotpotQA, HaluEval, and Mintaka contain such biases.

\begin{figure}[h!]
\begin{center}
\includegraphics[width=0.75\linewidth]{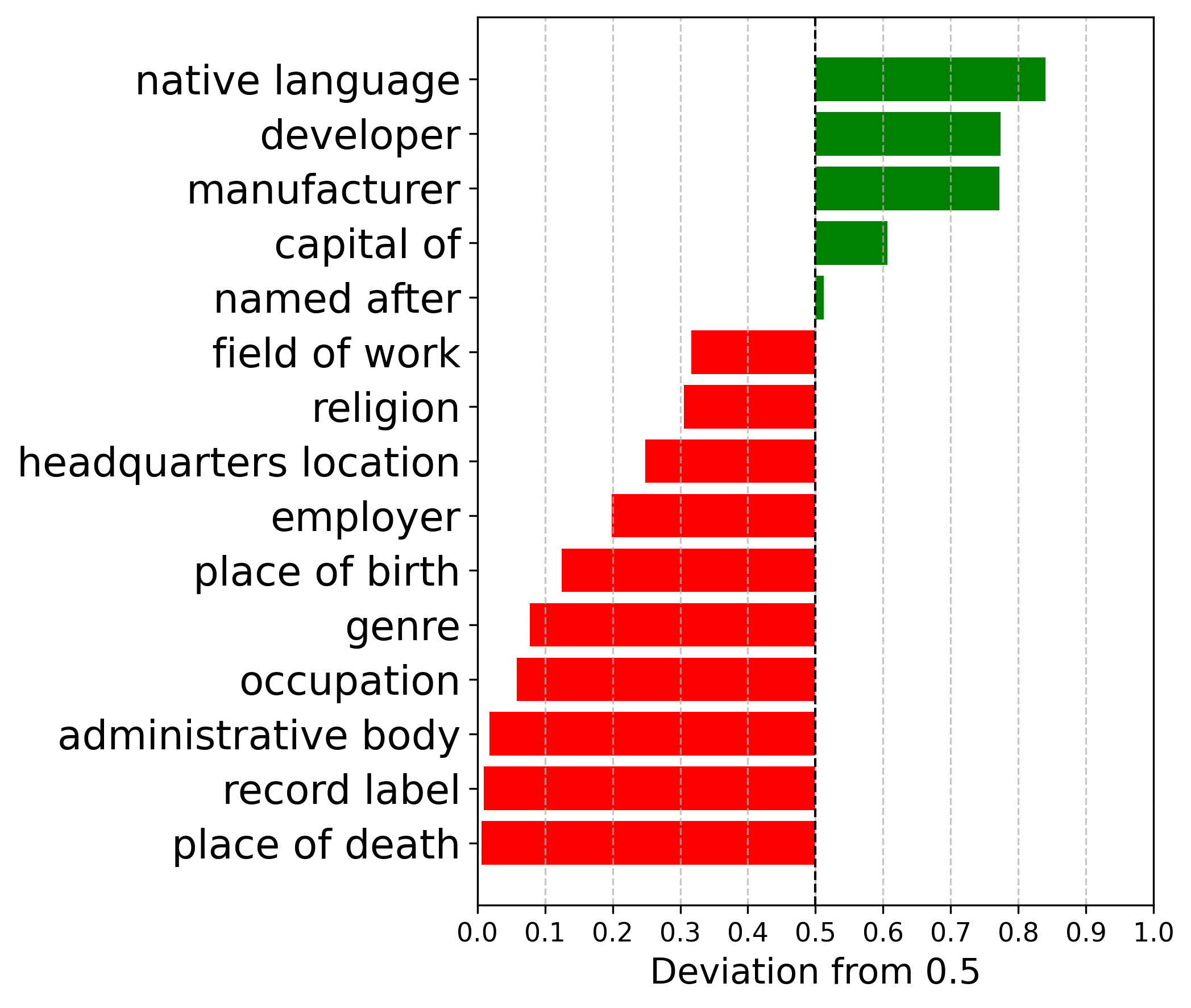}  
\end{center}
\vspace{-10pt}
\caption{The portion of $k=True$ for each domain, by \texttt{LLaMA-3-8B} model on the ParaRel train set. The rate is skewed toward 0 or 1 by domain, rather than being centered around 0.5.}
\label{fig:pararel_true_rate}
\end{figure}

\vspace{-5pt}

There are various other question-side shortcuts, which are described in the \autoref{app:case_study}. These shortcuts can be identified by considering various scenarios in which they may act as shortcuts, and it is likely that some shortcuts remain undiscovered, as it is very subtle to determine. Therefore, manually identifying and removing them from datasets is nontrivial. That is why we introduce AQE in the next section, a method for approximately assessing the total effect of question-side shortcuts without human investigation.

\subsection{Approximate question-side effect}
\label{sec:aqe}
In this section, we describe the concept of AQE.
Repeating \autoref{sec:definition}, the model $\theta$ is given a question $x$ to generate answer, where the correctness is denoted as a binary representation $k$. A hidden state $s$ is extracted at the first position of the answer. And $s$ is  consist of model-side $s_M$ and question-side information $s_Q$. Learnable module $\phi$ can perceive $s$ to infer $\hat{k}$ to predict $k$, which is hallucination prediction. Self-awareness is defined as utilizing $s_M$ to predict $k$. Here, what we ultimately aim to measure is the effect of utilizing $s_M$ in predicting $k$, denoted as \( \mathcal{A}(\phi(s_M)) \). $\mathcal{A}(\cdot)$ is a metric that measures the correctness of the predicted $\hat{k}$. As all we can measure is \( \mathcal{A}(\phi(s_Q, s_M)) \), we decompose this as follows: $\mathcal{A}(\phi(s_Q, s_M)) \approx \mathcal{A}(\phi(s_Q)) + \mathcal{A}(\phi(s_M))$
 This allows us to measure the effect of self-awareness as follows:  
 \begin{eqnarray} \label{eq:aqe2}
\mathcal{A}(\phi(s_M)) \approx \mathcal{A}(\phi(s_Q, s_M)) - \mathcal{A}(\phi(s_Q))
\vspace{-20pt}
\end{eqnarray}

\paragraph{AQE as a Shapley analysis}
We formalize AQE as a special case of Shapley analysis \citep{fryer2021shapley}, which evaluates the impact of individual factors on the outcome. Specifically, AQE corresponds to the concept of \textbf{marginal contribution}—a metric that quantifies the separate contribution of a single factor.

\vspace{-15pt}
\begin{eqnarray} \label{eq:shapley}
\Gamma_\beta(\alpha) = \Gamma(\alpha\cup \beta) - \Gamma(\beta)
\end{eqnarray}
\vspace{-20pt}

\autoref{eq:shapley} is the general formulation of marginal contribution. Here, $\alpha$ and $\beta$ represent individual components of a system. $\alpha \cup \beta$ represents a state in which these components are mixed together. $\Gamma(\cdot)$ represents the baseline performance metric (e.g., AUROC). $\Gamma_\beta(\alpha)$ quantifies the impact of removing $\beta$, thus isolating the contribution of $\alpha$. In our setting, $\alpha$ and $\beta$ correspond to the model-side and question-side information, respectively. $\alpha \cup \beta$ can be interpreted as the information contained in the hidden state of the model, which integrates both the model and question side information. $\Gamma(\alpha\cup \beta)$, $\Gamma(\beta)$, and $\Gamma_\beta(\alpha)$ correspond to $A(\phi(s))$, $A(\phi(s_Q))$, and $A(\phi(s_M))$, respectively.

\paragraph{Computing of self-awareness}
Computing \( A(\phi(s_Q)) \) is achieved by introducing a distinct model \( \theta' \) (where \( \theta' \ne \theta \)) which is optimized to only embed basic properties of the input question (e.g., domain or question type), \( s'_Q \approx s' = \theta'(x) \). A representative example of $\theta'$ is sBERT \citep{reimers2019sentencebertsentenceembeddingsusing}. sBERT is a very small model with only 22.7M parameters, but it is optimized to generate an embedding vector \( s' \) from input text $x$ (e.g., question). sBERT is known to capture high-level information from text as effectively as $\theta$ with a larger architecture (e.g., \texttt{LLaMA-3-8B}), as long as the target information is simple enough (e.g., domain classification). Therefore, while $s_Q'$ and $s_Q$ reside in different representational spaces of two distinct models, they are assumed to capture similar high-level information ($s_Q' \sim s_Q$).
 Conversely, due to its small size, we can assume $\theta'$ holds a very small amount of knowledge, which makes the knowledge distribution of $\theta'$ and $\theta$ independent ($s'_M$ and $s_M$ are independent). This makes $s'_M$ ignored when $\phi$ learning to predict $k$ (correctness of $\theta$) from $s'$ (hidden state from $\theta'$).

This results in \autoref{eq:aqe3}, where \( \phi \) and \( \phi' \) share architecture but are trained independently, as \( s_Q \) and \( s_Q' \) lie in different representational spaces.
\vspace{-5pt}
\begin{eqnarray} \label{eq:aqe3}
\mathcal{A}(\phi(s_Q)) \approx \mathcal{A}(\phi'(s'))
\end{eqnarray}
\vspace{-15pt}

\begin{table}
\caption{AQE assessment on datasets. Prediction of $k$ from \texttt{LLaMA-3-8B}, with $s'$ from sBERT.}
\label{tab:aqe_original}

\centering

\resizebox{\linewidth}{!}{ 
\begin{tabular}{llllllll}
\toprule
\multicolumn{1}{l}{} & \multicolumn{5}{c}{short-form} & \multicolumn{1}{c}{long-form} 
\\ \cmidrule(lr){2-6}  \cmidrule(lr){7-7}  
&      ParaRel    & Mintaka      & HaluEval           & HotpotQA        & \makecell{Simple\\ Question}  & Explain     \\ \hline \addlinespace
$p(k=True)$             &  54.31       & 55.01            & 37.51            & 32.71            & 19.08    & 31.63\\ 
$p(k=False)$           &  45.68       & 44.98             & 62.48             & 67.28   & 80.19      &  68.36   \\ 
AQE$_{acc}$       & 73.26   & 63.50   & 68.55      &   72.50       &  82.36 &  65.65 \\
AQE$_{auc}$           & 82.61         & 66.67         & 68.37      &   70.14    &  68.13   &  69.40   \\

\bottomrule
\end{tabular} 
}

\vspace{-10pt}
\end{table}

The resulting \( \mathcal{A}(\phi(s_Q)) \) of \autoref{eq:aqe3} is defined as AQE. To summarize, $\theta'$ predicts $k$ of $\theta$ without using information of $\theta$. Intuitively, a distinct model $\theta'$ predicts whether $\theta$ will succeed on a given question, using only the information from the question. As no information about $\theta$ is involved, this setup excludes self-awareness and reflects only question-awareness. Together with \autoref{eq:aqe2}, we can derive \( \mathcal{A}(\phi(s_M)) \), the component of self-awareness in the measured hallucination prediction performance:  $ \mathcal{A}(\phi(s_M)) \approx \mathcal{A}(\phi(s)) - \mathcal{A}(\phi'(s'))$

\begin{table*}[h!]
\centering
\caption{AQE score of dataset and \texttt{LLaMA-3-8B} model. The version (original, type, domain) with the lowest AQE within each dataset is highlighted in \textbf{bold}.}
\label{tab:aqe_refined}
\resizebox{0.8\textwidth}{!}{ \begin{tabular}{
  ccccccccccc
}
\toprule
\multicolumn{1}{l}{} & \multicolumn{3}{c}{Mintaka} & \multicolumn{2}{c}{HotpotQA} & \multicolumn{2}{c}{ParaRel} & \multicolumn{2}{c}{Explain} 
\\ \cmidrule(lr){2-4}  \cmidrule(lr){5-6}  \cmidrule(lr){7-8} \cmidrule(lr){9-10}
&      original    & +type      & type + domain           & original    & +type  &  original & +domain &original & +domain   \\ \hline \addlinespace
$p(k=True)$     &  55.01       &  49.71    &    53.07     & 42.33       &  29.12       & 54.31    & 60.45    & 31.63  & 39.83   \\ 
$p(k=False)$     &  44.98       &  50.28   &    46.92     & 62.48       &   70.87      & 45.68   &  39.54    &  57.66 & 60.16 \\ 
AQE$_{acc}$       & 63.50   & 59.81   &  \textbf{59.04}     &   68.55       &  \textbf{76.03}  &  73.26 & \textbf{55.09} & 65.65   & \textbf{61.21}  \\
AQE$_{auc}$    & 66.67   & 64.06   & \textbf{61.62}      &   68.37          &  \textbf{55.37}   &  82.61 & \textbf{57.55} & 69.40   & \textbf{61.89}  \\
\bottomrule
\end{tabular} }
\vspace{-10pt}
\end{table*}

However, this formulation of AQE holds only under the assumption that \( s \) is the hidden state format. When \( s \) is a confidence score, AQE cannot be directly applied because confidence score is a one-dimensional value, which is too saturated to embed high-level information of the question.

\vspace{-5pt}
\subsection{Measuring AQE across datasets}
\label{sec:exp_aqe}
In this section, we measure AQE across hallucination prediction datasets. The model \( \theta \) is \texttt{LLaMA-3-8B-Instruct}\footnote{https://huggingface.co/meta-llama/Meta-Llama-3-8B}, and we evaluate it mainly on short-form factoid datasets (e.g., ParaRel, Mintaka, Halueval, HotpotQA), and additionally on long-form factoid datasets (e.g., Explain). The details of each dataset are provided in the \autoref{app:dataset}. We use two metrics: AUROC and accuracy. AQE for each metric is denoted as AQE$_{auc}$ and AQE$_{acc}$, respectively. We also report the $p(k=True)$ and $p(k=False)$, the bias of binary label $k$. We show that a larger (70B) model shows similar trends (\autoref{app:additional_exp}).

As shown in \autoref{tab:aqe_original}, the AQE$_{auc}$ for most datasets is close to or exceeds 0.70, rather than centering around 0.5. This indicates that many datasets exhibit a strong effect of question-side shortcuts, and a model can easily achieve AUROC over 0.70 without any self-awareness (i.e., perception of model-side information), relying solely on question-aware skills such as domain classification. Though previous works \citep{snyder2024early, zhang2024rtuninginstructinglargelanguage} have reported to achieve AUROC over 0.80 in hallucination prediction using these datasets, it could be a statistical artifact that is hard to generalize. In all datasets, AQE$_{acc}$ is consistently higher than the $p(k=True)$, indicating that AQE captures question-side effect beyond just the bias of the binary label.

\subsection{AQE in refined dataset}
In \autoref{sec:case_study}, we analyze that the type and the domain of a question can act as question-side shortcuts in predicting $k$. Fortunately, some datasets (HotpotQA, Mintaka, and ParaRel) include tags for this information, which allows us to control for it. We analyze AQE for each dataset before and after this control. In \autoref{tab:aqe_refined}, the ``+type'' column refers to the dataset after excluding binary-type questions. ``+domain'' indicates a regrouping of the train and test data according to their categories, such that the domains do not overlap (i.e., out-of-domain setting). The refinement process is detailed in \autoref{app:dataset}.

The experimental results can be summarized in two main points.  
(1) Applying refinement leads to a significant reduction in AQE. This demonstrates that through post-processing, we can get a dataset with lower dependency on question-side information, which is more suitable for evaluating self-awareness. (2) AQE still remains even after refinement. This suggests the presence of a question-side effect that we have not yet identified or controlled for.

\subsection{SCAO}
\vspace{-5pt}
We propose a method called Semantic Compression by Answering in One word (SCAO), which leverages model-side information more effectively by instructing the model to “answer in one word,” thereby improving the alignment of confidence values. We provide further explanation in \autoref{app:scao}.


\section{Experiment on hallucination prediction approaches}
\label{sec:experiment}

In this section, we evaluate hallucination prediction approaches across multiple datasets and their refined versions.

\begin{table*}[h!]

    \caption{Hallucination prediction performance (AUROC) of instruction-tuned 8B and 70B LLaMA models across multiple datasets.}
    \label{tab:main_exp}
    \subcaption{Mintaka}
    \centering
    \resizebox{0.85\textwidth}{!}{ \begin{tabular}{ccccccccccccc} 
    \toprule
    \multicolumn{1}{c}{} & \multicolumn{6}{c}{8B} & \multicolumn{6}{c}{70B} \\ 
    \cmidrule(lr){2-7} \cmidrule(lr){8-13}  
    \multicolumn{1}{c}{} 
      & \multicolumn{2}{c}{original} 
      & \multicolumn{2}{c}{+ type} 
      & \multicolumn{2}{c}{+ type + domain} 
      & \multicolumn{2}{c}{original} 
      & \multicolumn{2}{c}{+ type} 
      & \multicolumn{2}{c}{+ type + domain} \\ 
    \cmidrule(lr){2-3}  \cmidrule(lr){4-5}  \cmidrule(lr){6-7}  
    \cmidrule(lr){8-9}  \cmidrule(lr){10-11}  \cmidrule(lr){12-13}
    & auroc & $\mathcal{A}(\phi(s_M))$ 
    & auroc & $\mathcal{A}(\phi(s_M))$ 
    & auroc & $\mathcal{A}(\phi(s_M))$ 
    & auroc & $\mathcal{A}(\phi(s_M))$ 
    & auroc & $\mathcal{A}(\phi(s_M))$ 
    & auroc & $\mathcal{A}(\phi(s_M))$ \\ 
    \midrule \addlinespace
    Conf      &   64.88        &      -        &   64.61         &  -    &   69.23         &    -   
 & 69.41 & - & 67.16 & - & 66.35 & - \\ 
    Conf (\textbf{SCAO}) &    72.13       &    -         &  72.01          &  -    &  \textbf{75.51}          &  - & 73.64 & -   & 72.78 & -  & \underline{72.46} & -   \\ 
    Probe$_{dnn}$  &  77.09   &  8.54       &    75.70     & 11.64        &  72.79    &  11.17  & 76.43 & 10.68 & 75.54  & 10.19 & 71.03 & 12.26   \\ 
    Conf + Probe   &    \underline{79.10}   &  10.55    & \underline{ 77.54}    & 13.48    &    70.77    &   9.15 & \textbf{78.93} & 13.18 &\textbf{ 77.68}  & 13.33 & 70.80 & 12.03   \\
      Conf + Probe (\textbf{SCAO})       &    \textbf{79.41}   &  10.86    &  \textbf{77.89}    & 13.83      &    \underline{74.89}    &   13.27 & \underline{78.86}  & 13.11 & \underline{77.22} & 12.87 & \textbf{72.89} & 14.12   \\
    \bottomrule
    \end{tabular} } 
\end{table*}

\begin{table}
\vspace{-15pt}
\subcaption{HotpotQA}
\resizebox{\linewidth}{!}{ 
\begin{tabular}{
  cccccccccccccccccccc
}
\toprule
\multicolumn{1}{l}{} & \multicolumn{4}{c}{8B}   & \multicolumn{4}{c}{70B} \\ \cmidrule(lr){2-5} \cmidrule(lr){6-9} 
\multicolumn{1}{l}{} & \multicolumn{2}{c}{original} & \multicolumn{2}{c}{+ type}   & \multicolumn{2}{c}{original} & \multicolumn{2}{c}{+ type}
\\ \cmidrule(lr){2-3}  \cmidrule(lr){4-5}  \cmidrule(lr){6-7}  \cmidrule(lr){8-9}  
               & auroc        &$\mathcal{A}(\phi(s_M))$    & auroc         & $\mathcal{A}(\phi(s_M))$   & auroc        & $\mathcal{A}(\phi(s_M))$   & auroc         & $\mathcal{A}(\phi(s_M))$       \\ \hline \addlinespace
Conf         &  74.88          &      -      & 68.82           &    -   & 73.33 & -  & 72.87  & -  \\ 
Conf (\textbf{SCAO})     &   77.70       &       -        &    73.81        &    -  & 74.13 & -  & \underline{73.42}  & -   \\
Probe$_{dnn}$   &   80.58   &  12.21    &    73.17     & 17.80   &  74.41 & 10.63 & 69.42 & 14.86    \\ 
Conf + Probe   &    \underline{81.08}   &  12.71   &  \underline{73.87}    & 18.50   & \textbf{77.33} & 13.55  & 73.06 & 18.50 \\
   Conf + Probe (\textbf{SCAO})      &    \textbf{83.39}   &  15.02   &  \textbf{75.51}    & 20.14  & \underline{77.28} & 13.50 & \textbf{73.52} & 18.96    \\
\bottomrule
\end{tabular} 
} 
\subcaption{ParaRel}
\resizebox{\linewidth}{!}{ \begin{tabular}{
  cccccccccccccccccccc
}
\toprule
\multicolumn{1}{l}{} & \multicolumn{4}{c}{8B}   & \multicolumn{4}{c}{70B} \\ \cmidrule(lr){2-5} \cmidrule(lr){6-9} 
\multicolumn{1}{l}{} & \multicolumn{2}{c}{original} & \multicolumn{2}{c}{+ domain}   & \multicolumn{2}{c}{original} & \multicolumn{2}{c}{+ domain}
\\ \cmidrule(lr){2-3}  \cmidrule(lr){4-5}  \cmidrule(lr){6-7}  \cmidrule(lr){8-9}  
               & auroc        &$\mathcal{A}(\phi(s_M))$    & auroc         & $\mathcal{A}(\phi(s_M))$   & auroc        & $\mathcal{A}(\phi(s_M))$   & auroc         & $\mathcal{A}(\phi(s_M))$       \\ \hline \addlinespace
Conf        &    71.03       &    -        &    59.51        &      -   &  70.87  & - & 59.92 \\ 
Conf (\textbf{SCAO})    &    69.23       &       -    &   \underline{73.12}         &   -     & 73.45  & - & \textbf{74.51}  \\ 
Probe$_{dnn}$    &   88.76   &  6.15       &    70.34     & 12.79     & \underline{89.88} & 2.90 & 68.66 & 15.37   \\ 
Conf + Probe    &    \underline{88.78}   &  6.17    &  73.08    & 15.53   & \textbf{90.01 }  & 3.03 & 68.86 & 15.57  \\
Conf + Probe (\textbf{SCAO})       &    \textbf{88.95}   &  6.34   &  \textbf{76.09}    & 18.54  & 89.82  & 2.84 & \underline{70.84} & 17.55 \\
\bottomrule
\end{tabular} } 
\subcaption{Explain}
\resizebox{\linewidth}{!}{ \begin{tabular}{
  ccccccccccccc
}
\toprule
\multicolumn{1}{l}{} & \multicolumn{4}{c}{8B}   & \multicolumn{4}{c}{70B} \\ \cmidrule(lr){2-5} \cmidrule(lr){6-9} 
\multicolumn{1}{l}{} & \multicolumn{2}{c}{original} & \multicolumn{2}{c}{+ domain}   & \multicolumn{2}{c}{original} & \multicolumn{2}{c}{+ domain}
\\ \cmidrule(lr){2-3}  \cmidrule(lr){4-5}  \cmidrule(lr){6-7}  \cmidrule(lr){8-9}  
               & auroc        &$\mathcal{A}(\phi(s_M))$    & auroc         & $\mathcal{A}(\phi(s_M))$   & auroc        & $\mathcal{A}(\phi(s_M))$   & auroc         & $\mathcal{A}(\phi(s_M))$       \\ \hline \addlinespace
Conf       &  49.45   &     -   &    50.57 &    -   & 48.96   & - & 46.81   & -  \\ 
Conf (\textbf{SCAO})    & 62.90   &    -  &   62.28         &   -   & 57.21   & - & 59.54 & -    \\ 
Probe$_{dnn}$   &   84.68   &  15.28  &    68.55     &  6.66  & \underline{83.69}  & 16.00 & 66.58 & 8.87  \\ 
Conf + Probe    &    \underline{84.89}  &  16.49  &   \underline{69.15}   &  7.26   & 83.68  & 15.99 & \underline{66.78} & 9.07 \\
 Conf + Probe (\textbf{SCAO})      &   \textbf{85.42}   &  17.02    & \textbf{70.04}    & 8.15  & \textbf{84.94}  & 17.25 & \textbf{68.67} & 10.96  \\
\bottomrule
\end{tabular} } 
\end{table}


\textbf{Approaches  \ \ \  }
Previous approaches for hallucination prediction can be broadly categorized into three. 
(1) confidence-based: this utilizes the softmax probability of the answer token \citep{fadeeva2024fact}. It is utilized in other forms, such as perplexity \citep{ren2023outofdistributiondetectionselectivegeneration} or energy \citep{liu2021energybasedoutofdistributiondetection}. We adopt a simplified method that takes the mean of top-\( n \) softmax probabilities of the first answer token and applying a threshold. \( n \) and the threshold \( t \) are learnable $\phi$. 
(2) hidden-state-based: This approach feeds the hidden-state vectors of a model $\theta$ into learnable $\phi$. $\phi$ can be a linear layer \citep{li2024inference} or more complex architecture \citep{azaria2023internal, chen2024insidellmsinternalstates}. We adopt a three-layer deep neural network. (3) aggregation: This approach concatenates the confidence scores and hidden state into a single vector, which is then passed to a learnable \( \phi \) \citep{snyder2024early}.
In \autoref{tab:main_exp}, \textit{Conf}, \textit{Probe}, and \textit{Conf + Probe} represent the three evaluated approaches, respectively. Detailed explanations are provided in \autoref{app:three_appoach}. We conduct experiments using instruction-tuned LLaMA models of two different sizes (8B, 70B).

\textbf{Dataset and metric \  \ \ }
For the evaluation of hallucination prediction, we narrow down our focus to the datasets from \autoref{sec:exp_aqe} that support refinement (Mintaka, HotpotQA, ParaRel, Explain). We also report the gap between the metric performance and AQE, denoted as $\mathcal{A}(\phi(s_M))$. However, we do not report the $\mathcal{A}(\phi(s_M))$ for  the methods that use only the confidence score (e.g., \textit{Conf}), because they lack question-side information, making it impossible to compute \( s_Q \). In \autoref{tab:main_exp}, \textit{original} refers to the unrefined version of dataset, while \textit{+ type} and \textit{+ domain} indicate versions refined based on question type and domain, respectively. Detailed descriptions are in \autoref{app:dataset}.

\textbf{Results \ \ \ }
We observe the following points.
(1) On the datasets of \textit{original}, the performance shows very promising records of around AUROC 0.80, the highest among all versions. However, the $\mathcal{A}(\phi(s_M))$ is the smallest, suggesting that the performance is largely attributed to question-side shortcuts. (2) In refined versions (\textit{+ type, + domain}), the performance measures sharply drop. For example, the AUROC for HotpotQA drops from 80.58 to 73.17 after refinement. This again demonstrates that the performance reported in previous works was largely driven by shortcuts.

(3) However, the AQE gap is larger in the refined datasets, indicating that when question-side effects are reduced, the use of model-side information becomes more activated. (4) Methods that rely solely on the confidence score (\textit{Conf(SCAO)}) perform poorly on the original datasets, but show smaller performance variation across different data versions. And in some refined settings, it even outperforms other baselines. It is somewhat counter-intuitive that \textit{Conf(SCAO)} outperforms hidden-state based methods, though it is provided significantly smaller amount of information. This suggests that the confidence score with SCAO captures a substantial amount of \( s_M \), contributing to its strong generalization performance. (5) The \textit{Conf + Probe (SCAO)} shows the largest $\mathcal{A}(\phi(s_M))$ across all refined versions of the datasets, suggesting a stable and effective direction for achieving self-awareness. (6) These trends remain consistent across models of different sizes and also hold under the accuracy metric \autoref{app:exp_acc}.

(7) The $\mathcal{A}(\phi(s_M))$ is very low in OOD setting of Explain, which reveals that the hidden-state-based approach has limited generalization in long-form question answering settings. such as Explain. This contradicts the reports from previous works \citep{snyder2024early, chen2024insidellmsinternalstates}.

\section{Conclusion}


In this work, we argued that hallucination prediction performance is often inflated by question-side shortcuts rather than reflecting genuine self-awareness. To disentangle these effects, we introduced Approximate Question-side Effect (AQE) and showed that many existing benchmarks exhibit strong question-aware signals, limiting the generalizability of prior results. Our analysis demonstrates that, once such effects are controlled, the contribution of true model-side awareness is substantially smaller than previously reported. Limitations are discussed in \autoref{app:limitation}.



\newpage
{\small
\bibliography{main}
}

\newpage
\appendix

\section{Limitation}
\label{app:limitation}
\paragraph{Scope Narrowed to System 1}
As discussed in the \autoref{app:related}, our study focuses on System 1 (fast, automatic processing) rather than System 2. The knowledge recall setting we consider largely involves rapidly retrieving stored information in response to a prompt; accordingly, our analyses and conclusions should be interpreted primarily as describing phenomena that arise in retrieval/recall-driven, short-horizon judgments and answer generation.

In contrast, tasks dominated by System 2 (slow, deliberative processing)—such as multi-step reasoning, planning, long-form generation, and explicit verification—may exhibit qualitatively different sources and signals of failure. Therefore, the indicators and observations proposed in this work should not be assumed to directly generalize to System 2–heavy settings.

That said, this scope choice does not diminish the significance of our contribution. Because System 1 and System 2 reflect fundamentally different computational regimes and error profiles, concepts like self-awareness and hallucination detection are likely to carry distinct meanings and objectives in each regime. Our work provides a foundation for more precise definition, measurement, and analysis of these concepts in System 1 settings, and clarifies what may need to be preserved versus redesigned when extending to System 2 scenarios.

\paragraph{Devising robust methods}
How to design a more generalizable methodology remains an open question for future work.
Although we propose SCAO as a methodology, it still exhibits a very high AQE (i.e., low genuine awareness) in long-form question answering. This indicates that hallucination prediction results for long-form QA should be interpreted with great caution when they are based on existing confidence- or hidden-state–based methods.

Moreover, this suggests that long-form answering involves more complex functions beyond simple knowledge recall, and therefore requires hallucination prediction approaches that go beyond solely leveraging a model’s internal states.

\section{Other question-side shortcuts}
\label{app:case_study}

\textbf{ Broken question \  } 
The most commonly observed problem across all datasets is insufficient annotations of question and answer pairs. This occurs when questions and answers follow a one-to-many relationship, but the annotations fail to cover them. For example, the ParaRel consists of question-answer pairs such as: “Q: What field does Daniel Bernoulli work in? A: physics”. Although Daniel Bernoulli also worked in the fields of mathematics and medicine (one-to-many relation), only one label is provided, failing to cover all possible correct answers. If an LLM responds with a correct answer but different from the given label (e.g., mathematics), it would still be classified as hallucinated (i.e., $k$ is annotated $False$), which is incorrect. If \(\phi\) is trained to predict \( k \) using such \( \{s, k\} \) pairs, it will likely become biased toward predicting \( \hat{k} = False \) whenever the domain of question is ``field of work''. This means $\phi$ learns a domain classification task, not self-awareness. While this may improve prediction performance within the dataset, it lacks generalizability. It is because its performance is likely to drop only if the quality of the dataset improves, or if questions are given from unseen domains. As a study of \citet{zhang2024rtuninginstructinglargelanguage} reports hallucination performance on this dataset, this shortcut might have influenced the reported scores. This issue is also found in other datasets such as SimpleQuestions, which includes instances like ``Q: What is a TV action show? A: Genji Tsushin Agedama'', that also fail to cover various possible answers.

The broken question problem is addressed when a question includes detailed constraints that restrict the one-to-many mapping between the question and possible answers. For example, in Mintaka, detailed constraints are added to the questions to ensure a one-to-one mapping (e.g., ``Who was the director of The Goodfellas and attended school at New York University's School of Film?'').

\textbf{Domain \ \ } 
Classifying the domain of a question (e.g., science, society) alone can provide a rough guess of $k$. For example, suppose a model that is extensively trained on science data but is unfamiliar with society or history domains. Since the model is likely to make more hallucination when given questions in the history domain. In this way, $k$ can be biased toward $True$ or $False$ by domains, as shown in \autoref{fig:pararel_true_rate}. In such cases, the task of predicting \( k \) with \( \phi \) is more a domain classification, which is question-aware and not model-aware.

When we consider a case of a human, it becomes clear that such shortcuts have limited effectiveness and are far from self-aware. For example, suppose this model is not familiar with the history domain but still has knowledge about Abraham Lincoln. If this model were human, he could easily ``feel'' his inner state and recognize that he possesses knowledge about Lincoln, despite not being familiar with other historical issues. However, if the model were an LLM and $\phi$ is trained to exploit only question-side shortcuts, $\phi$ would tell the model does not know about Lincoln, though the model actually possesses the knowledge of Lincoln. Therefore, while the domain information of a question can provide a naive approximation of $k$, relying on it imposes limitations on precision. This again highlights that utilizing self-awareness is the ultimate direction in precise hallucination prediction.

\begin{table}
\caption{The portion of correct answer for each question type, by \texttt{LLaMA-3-8B}. The rate for binary type is in \textbf{bold}.}
\label{tab:answering_type}

\begin{minipage}{0.45\linewidth}

\subcaption{HotpotQA}

\resizebox{\linewidth}{!}{
\centering
\begin{tabular}{cc}
\hline
Type &$p(k=True)$\\ \hline
Bridge &  0.6828 \\
Comparison & \textbf{0.8477} \\
\bottomrule
\end{tabular} } 
\end{minipage}
\begin{minipage}{0.5\linewidth}
\subcaption{Mintaka}
\resizebox{\linewidth}{!}{
\centering
\begin{tabular}{cc}
\hline
Type & $p(k=True)$\\ \hline
Entity & 0.5508 \\
numerical & 0.4011 \\
date & 0.5968 \\
string & 0.6315 \\
boolean & \textbf{0.7283} \\
\bottomrule
\end{tabular} }
\end{minipage}
\end{table}

\textbf{Question type \ \ } 
Question type (e.g., short-answer, multiple-choice) also provides a strong hint for predicting \( k \). The average probability of $k=True$ (denoted as $p(k=True)$) for binary-choice questions is at least 0.5 for random choice, significantly higher than for open-ended questions with 0 for random choice, as described in \autoref{tab:answering_type}. In such a case, \(\phi\) may learn a shortcut where it automatically predicts \( k = True \) whenever it detects a binary-choice question. HotpotQA, HaluEval, and Mintaka contain such binary-choice questions.

There are various other question-side shortcuts, which are described in the \autoref{app:case_study}. These shortcuts can be identified by considering various scenarios in which they may act as shortcuts, and it is likely that some shortcuts remain undiscovered, as it is very subtle to determine. Therefore, manually identifying and removing them from datasets is nontrivial. That is why we introduce AQE in the next section, a method for approximately assessing the total effect of question-side shortcuts without human investigation.

\paragraph{Answerability  } 
Some questions are inherently unjudgeable in terms of correct or incorrect. This includes preference-based questions, hypothetical scenarios, and philosophical inquiries. For such questions, any answer could be considered correct or hallucinated, depending on the perspective of the evaluator. For example, if an LLM is asked ``What color do you like?'' and responds ``Blue,'' the correctness of this answer depends on the interpretation: 1) If correctness is judged based on contextual appropriateness, the answer is valid. 2) If correctness is judged based on the idea that LLMs cannot have personal preferences, the answer could be labeled as hallucinated. In this case, \(\phi\) can exploit this pattern by simply identifying unanswerable questions and assigning a fixed label (either \( k = 0 \) or \( k = 1 \)) across all such cases. This makes the hallucination detection process dependent on question-awareness, rather than assessing self-awareness.

The dataset SelfAware collects only unanswerable questions and categorizes them into different types (e.g., no scientific consensus, imagination, completely subjective, too many variables, philosophical). However, distinguishing these does not require self-awareness. Instead, it is primarily an act of reading comprehension, where the model identifies the nature of the question rather than assessing its own knowledge state.

\paragraph{Time-sensitive question } Time-sensitive questions are inherently difficult for LLMs to answer accurately, as LLMs lack a robust understanding of time \citep{jain-etal-2023-language-models}. As a result, questions involving temporal information will be biased toward hallucinated responses. Datasets such as HaluEval, Mintaka, and TruthfulQA include such questions (e.g., "How old is Barack Obama?", "When did the most recent pandemic occur?").

\paragraph{Complexity } Complexity awareness is another question-dependent approach to estimating \( k \). This aligns with the case of ``too many variables'' in answerability awareness, where, if a question is too difficult, the model is more likely to fail, making it advantageous to predict "unknown" by default. However, the notion of complexity is relative. If a model has extensive knowledge in a certain domain, it may still answer correctly even if the complexity is high. These attributes may connect complexity with category awareness. Additionally, our analysis suggests that questions within a single dataset tend to have similar levels of complexity. Therefore, distinguishing \( k \) based on complexity within a dataset is expected to be relatively rare in the general experimental setup.
\section{Details on SCAO}
\label{app:scao}

In this section, we propose Semantic Compression through Answering in One word (SCAO), a hallucination prediction method designed to maximize the utilization of \( s_M \). While the \(\phi\) predicts \(k\) from input \(s\), which consists of both \(s_M\) and \(s_Q\), \(\phi\) tends to depend more on \(s_Q\) which offers an easier shortcut, as discussed in \autoref{sec:case_study}. To prevent this problem, we need a method that strengthens the preference of $\phi$ for \( s_M \).

To address this, we focus on the confidence score as a source of $s$. In \autoref{sec:aqe} we noted that the confidence score is a 1-dimensional scalar where information is extremely saturated, which in turn makes it unlikely to carry high-level information of the question ($s_Q$). This makes the confidence score closer to \( s_M \) ($s\approx s_M$). Therefore, using the confidence score alone or aggregating it with the hidden state may increase the model’s dependency on \( s_M \). 

\begin{figure*}
\centering
  \includegraphics[width=0.8\textwidth]{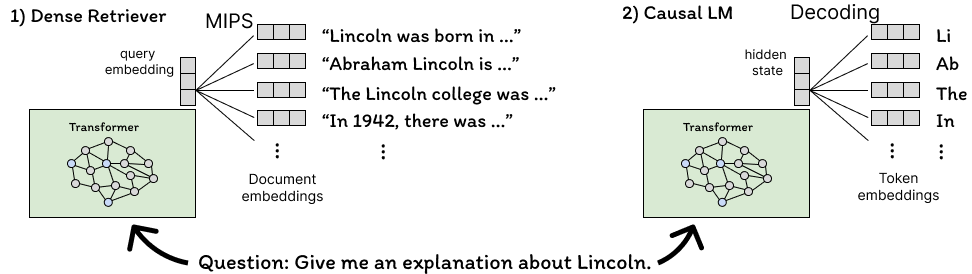}
  \caption{Structural analogy between 1) dense retriever and 2) causal LM.}
  \label{fig:retriever_compare}
\end{figure*}

However, as the confidence score is highly saturated and carries limited information, an approach is needed to amplify the information contained in it. SCAO is a method designed to more effectively express confidence in knowledge that can be represented as entities. The approach is straightforward: we insert a system instruction before the question, prompting the model to ``you must answer in one word''. The rationale behind why this technique can improve the use of $s_M$ is as follows.

\textbf{Causal LLM is an entity retriever when compressed}
The confidence score is calculated by the maximum inner product score between the last hidden state of $\theta$ and the token embedding vectors (i.e., decoder head). We analyze that this structure is analogous to the maximum inner product search (MIPS) used in dense retrieval \citep{karpukhin2020dense} (\autoref{fig:retriever_compare}). And we focus on the calibration function of similarity scores in dense retrieval. Previous research on dense retriever systems such as Faiss \citep{douze2024faiss} leverages this calibration through a \textit{range search}, which finds all the document vectors that are within some distance threshold. 

This concept can be interpreted in reverse that we can evaluate whether knowledge is within the vector DB, with a fixed confidence threshold. For example, suppose that a vector database contains documents about the biography of Newton but contains rare data about the biography of Lincoln. Querying ``Give me an explanation on Lincoln'' may result in few documents with confidence scores over the threshold. In contrast, querying "Give me an explanation on Newton" would likely retrieve a greater number of documents exceeding the threshold, reflecting a stronger alignment between the query and the knowledge contained in the database.

\begin{figure}[!h]
  \centerline{
 \begin{subfigure}{0.46\linewidth} 
    \centering
    \includegraphics[width=\linewidth]{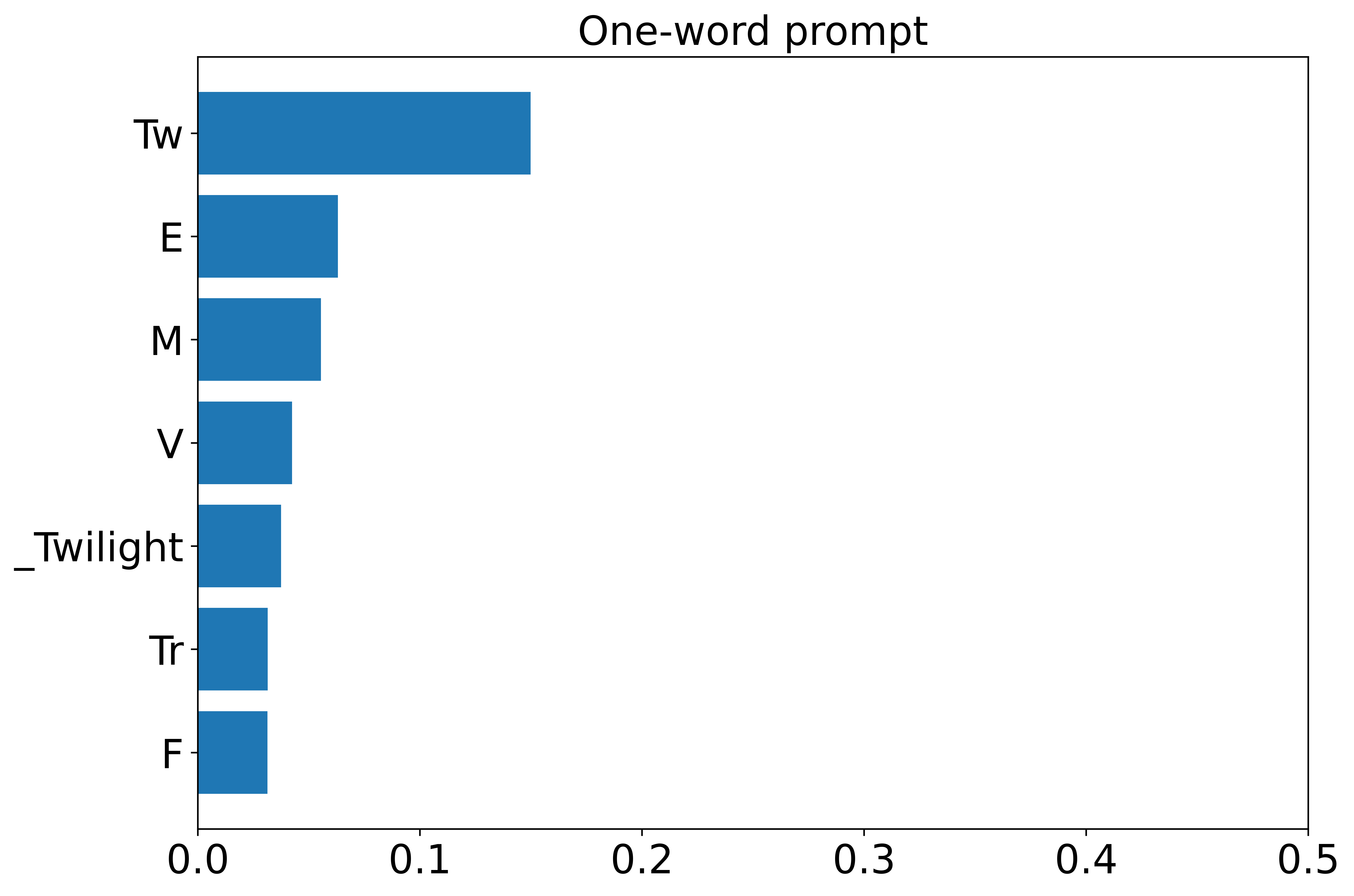}
    \end{subfigure}%
 \begin{subfigure}{0.46\linewidth} 
    \centering
    \includegraphics[width=\linewidth]{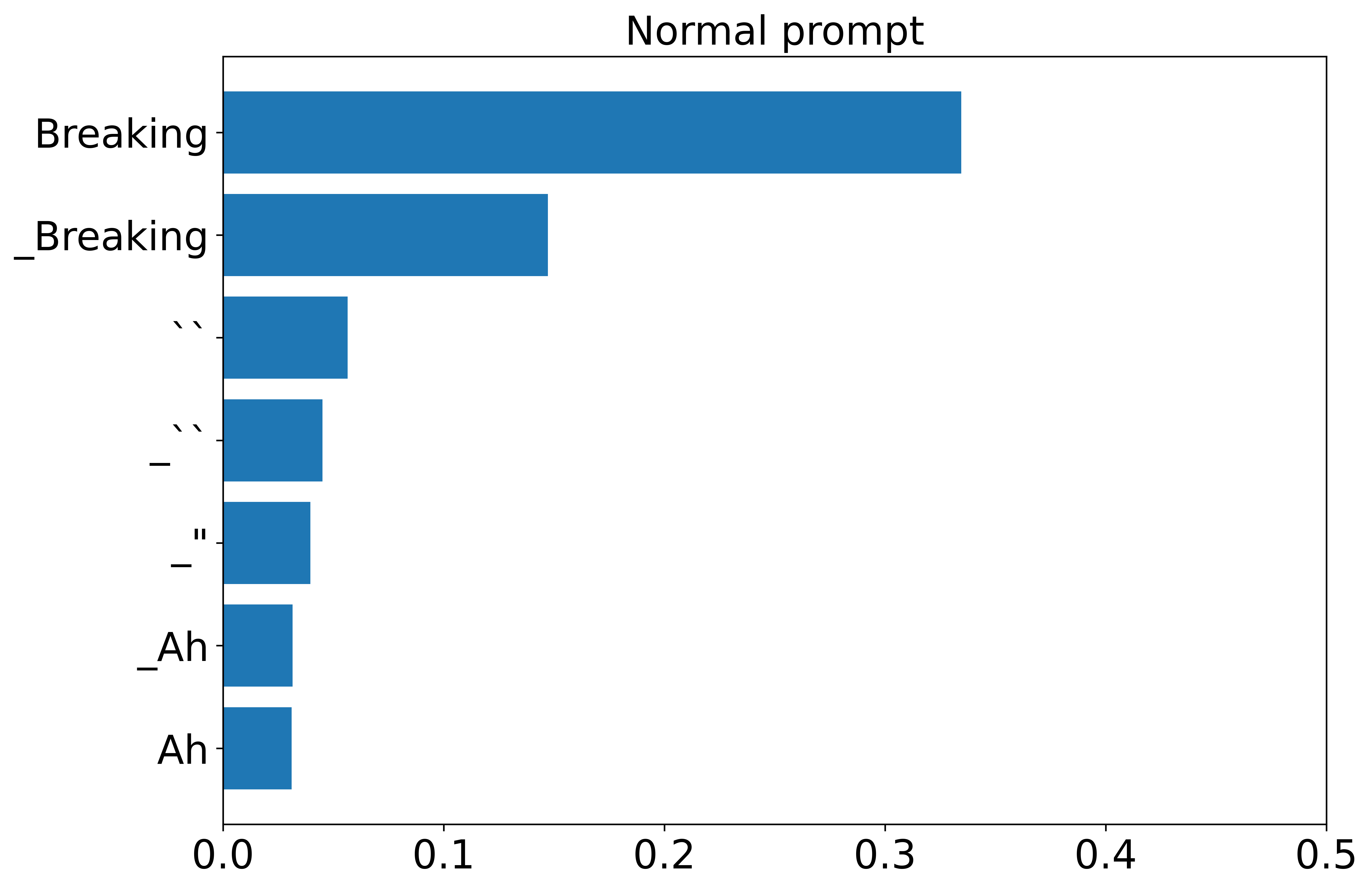}
    \end{subfigure}%
    }
  \caption{Y-axis is the top-7 candidates of the first token of the answer to the question ``Please give me an explanation about \textbf{Breaking Dawn}''. The X-axis is the probability for each candidate. \textbf{Left} is for one-word prompt, and the \textbf{Right} is for normal prompt.}
  \label{fig:open-ended-oneword}
\end{figure}

However, a causal LM not only performs entity retrieval but also generates full sentences by connecting these words. Such considerations of grammatical words and sentence structure may act as noise for a calibration signal. Therefore, by minimizing the consideration of sentence structure in the model, the LLM will become more analogous to an entity retriever. If LLM becomes more analogous to an entity retriever, its behavior will become more similar to the calibration properties in dense retrieval. A straightforward way to minimize the consideration of grammatical context in the model is to instruct it to ``you must answer in one word'', under the assumption that the model is well-trained to follow instructions. 

In \autoref{fig:open-ended-oneword}, we show the confidence pattern at the first token of the answer, with and without SCAO when the model is provided a question ``Please give me an explanation about Breaking Dawn''. With the one-word prompt (Left), the model appears to attempt to retrieve knowledge related to "Twilight," which is the series name of Breaking Dawn. In contrast, with the normal prompt (Right), the model tends to repeat the question entity, "Breaking Down". Since it chooses the easier path, the overall probabilities are higher. Further analysis of the confidence pattern is in \autoref{app:conf_pattern}. In \autoref{sec:experiment}, we empirically show that applying SCAO leads to improved hallucination prediction performance, especially more in settings with lower AQE dataset, where the use of model-side information becomes more critical.

\subsection{Efficiency of first token as a discriminator}
\begin{figure}
\includegraphics[width=0.9\linewidth]{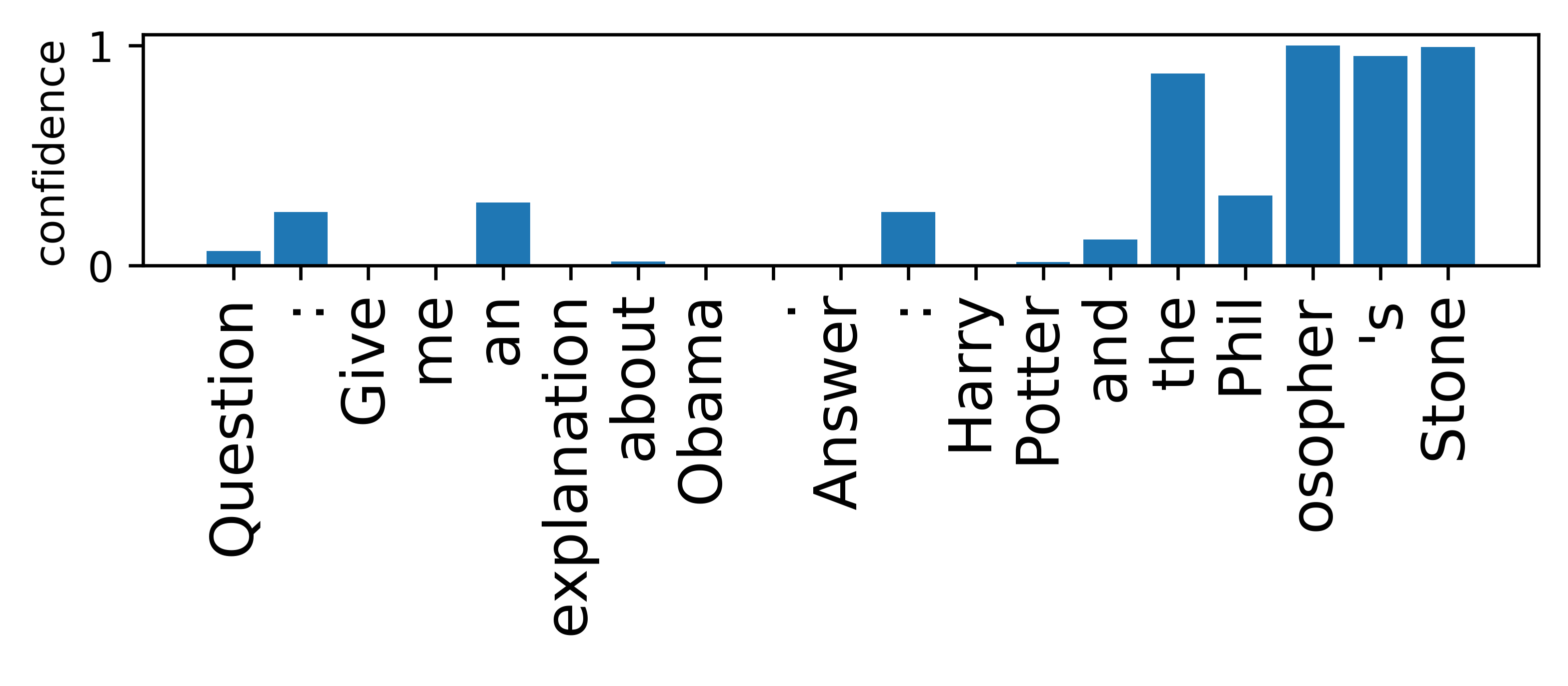}  
\caption{Probability pattern of the hallucinated answer, by LLaMA3-8B. Each bar stands for the probability (0,1) of the corresponding token.}
\label{fig:harry}
\end{figure}

Previous works on confidence-based hallucination detection research mostly utilize the confidence score of all tokens in the answer sentences, with normalization such as averaging \citep{chen2024insidellmsinternalstates}. 
Utilizing more information is ultimately more advantageous; however, it also has several drawbacks. We observe a pattern that as the entity name length increases, the average confidence tends to rise. For example, \autoref{fig:harry} depicts the confidence pattern of the hallucinated question-answer pair ``Question: Give me an explanation about Obama. Answer: Harry Potter and the Philosopher's Stone''.

\begin{figure}[!h]
    \centering
    \includegraphics[width=\linewidth]{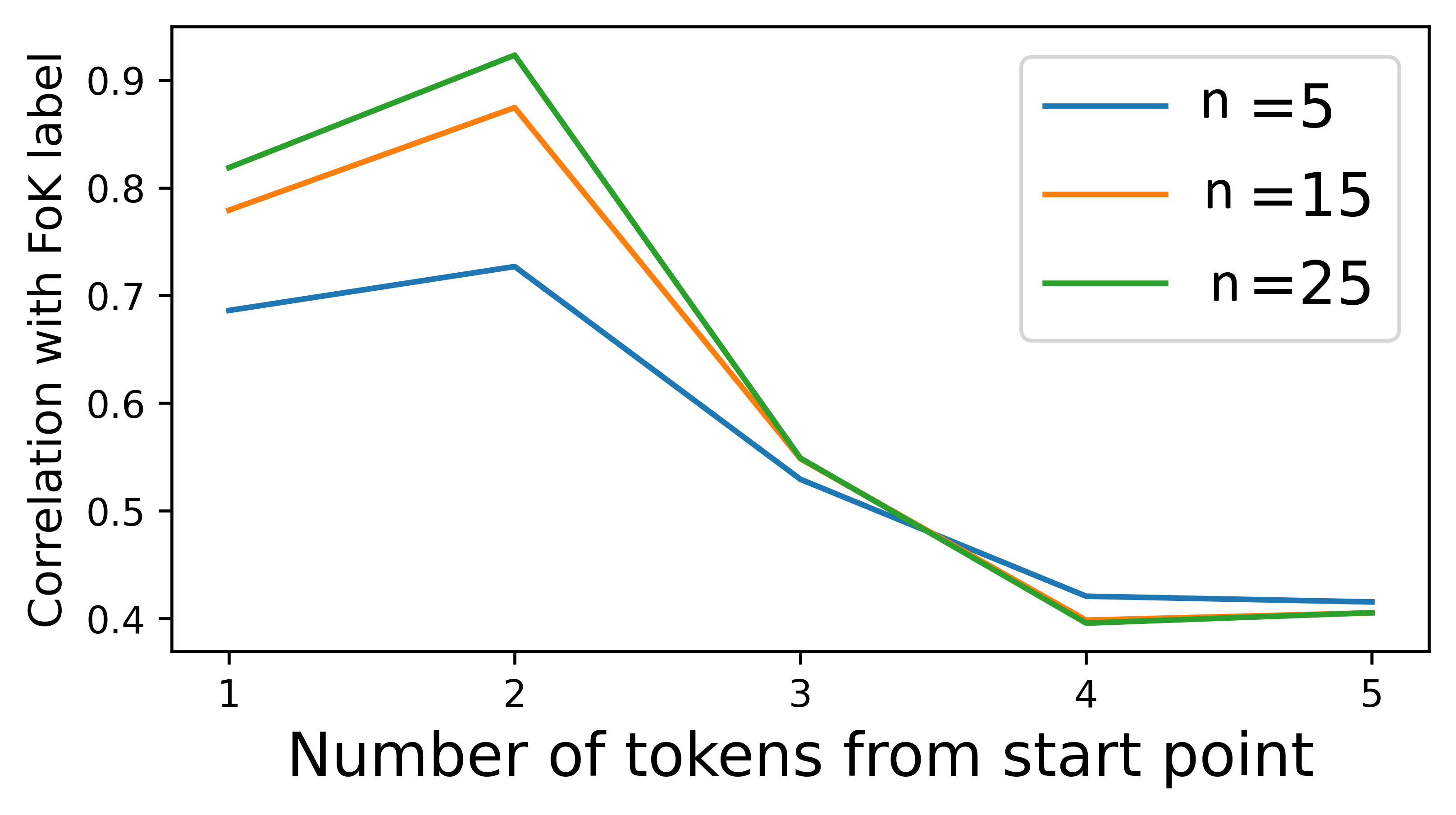}
    \centering
    \includegraphics[width=\linewidth]{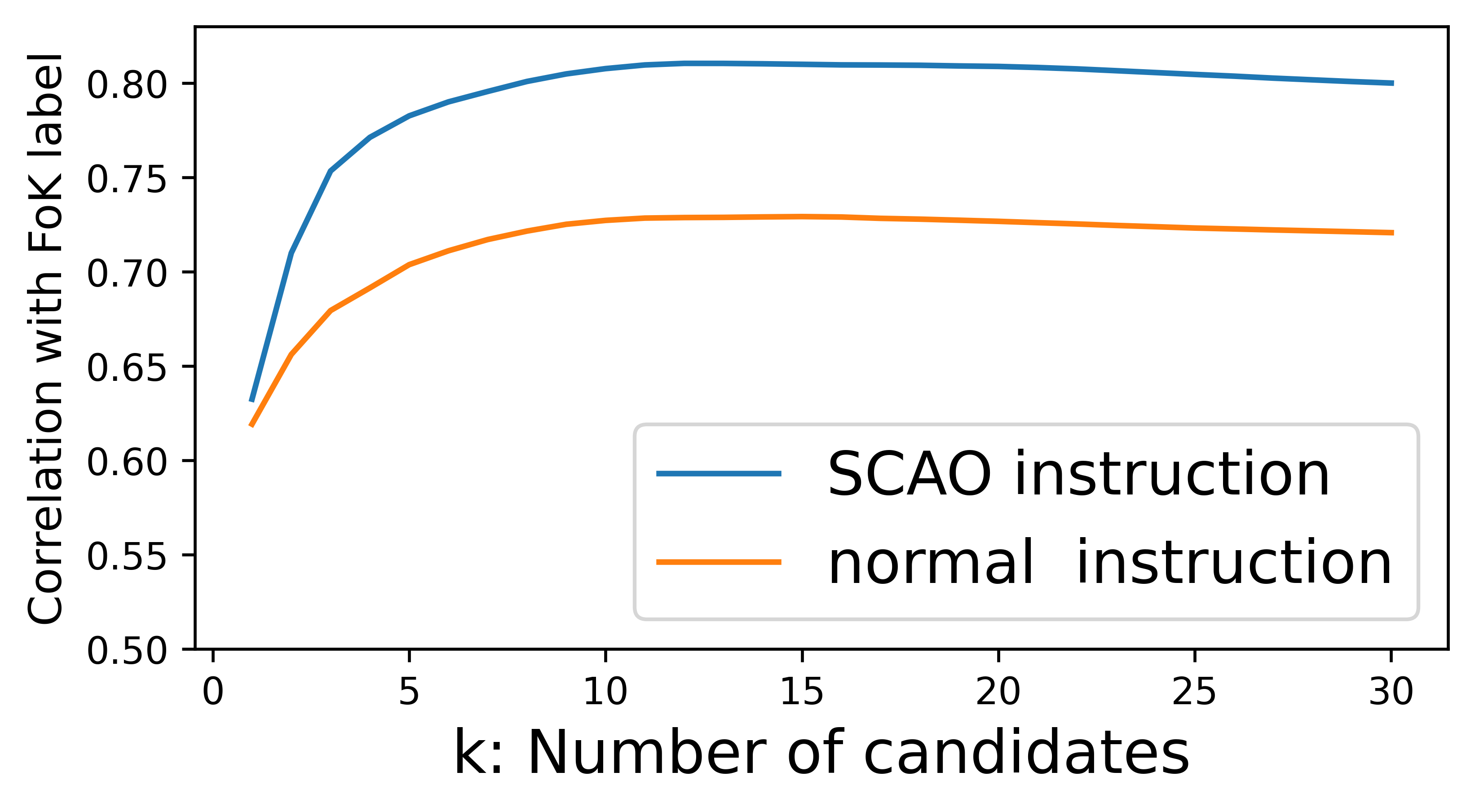}
  \caption{Y-axis is a correlation between the mean confidence and the $k$. The X-axis of each figure stands for (\textbf{Left}) the number of tokens from the start point of the answer, and (\textbf{Right}) the number of candidates used to calculate the mean. The LLaMA3-8B and ${(s,k)}$ datasets from Mintaka are utilized.}
  \label{fig:oneword_someword_k}
\end{figure}

Up to the token ``Harry Potter'', the confidence is near zero since it conflicts with the question. However, from a ``philosopher'', confidence increases to a near maximum, as the previous context of ``Harry Potter'' supports it strongly. Thus, the average confidence tends to increase regardless of whether it makes sense, when the entity name gets longer or the sentence contains more grammatical elements. This observation is supported by the analysis in \autoref{fig:oneword_someword_k} (Left), which shows that the correlation between the mean confidence and the $k$ tends to decrease as the token increases.

We also observe that averaging the confidence scores across top-$n$ vocabulary candidates, rather than just the top-1, shows a stronger correlation with the label $k$, particularly peaking around $n$=15 (\autoref{fig:oneword_someword_k} (Right)). This suggests that incorporating more samples of distance provides more information about the relationship between the output vector and the token space. 

\subsection{Confidence pattern of SCAO}
\label{app:conf_pattern}

\begin{table}[h!]

\centering
\caption{The number and portion of each case, when questions from the test set (total 2152) of Explain are asked to the \texttt{LLaMA-3-8B-Instruct} model using various prompts. The columns represent each prompt style. In the rows, ``repeating subject'' refers to cases where the top-1 candidate for the first token of the answer is a component of the queried subject entity. ``The'' refers to cases where the top-1 token is "the."}
\label{tab:open-ended-oneword}
\resizebox{\linewidth}{!}{
\begin{tabular}{
 lllllll
}
\toprule
                   & one-word prompt        & normal prompt \\ \hline 
repeating subject          & 1819 (84.5\%)      & 261 (12.1\%)      \\
"the"          & 383 (17.7\%)    & 5 (0.2\%)      \\ \hline
\end{tabular}
}
\end{table}
In \autoref{fig:open-ended-oneword}, we show how the model reacts at the first token of the answer in both one-word prompts and the normal prompt.

First, in non-compressed cases (queried with a normal prompt), the following patterns are frequently observed: (1) The response often starts by repeating the entity name mentioned in the query. (2) The response begins with grammatical function words such as "The" or "A". In other words, the model tends to take the easy path. As a result, the probability of the initial token is generally inflated, regardless of whether the model truly knows the subject.

On the other hand, when prompted to answer with a one-word response, the first token often corresponds to the initial token of a word encapsulating the entity's characteristics. For example, in response to the question "Please give me an explanation about 'Breaking Dawn'.", the first candidate token was "Tw" (the first token of "Twilight"). In other words, with one-word prompting, the model shows a stronger tendency to retrieve its own knowledge related to the entity.
This trend is also reflected statistically. Among the 2152 test samples in the Explain dataset, the case where the top-1 candidate of the first token of the response is a component of the entity is 84.5\% for normal prompting, significantly outpacing the 12.1\% for one-word prompting. Similarly, the first token being "the" occurred in 17.8\% of normal prompting cases, compared to just 0.02\% for one-word prompting. (\autoref{tab:open-ended-oneword})

\subsection{Rationale on why confidence-based method is better in generalization}
\label{app:scao_probe}

In \autoref{sec:experiment}, we observed that the confidence-based method (\textit{SCAO}) outperforms the hidden-state-based method (probing) in the out-of-domain setting. This result is counter-intuitive, as confidence scores are highly saturated scalar values, whereas hidden states are high-dimensional vectors capable of carrying much richer information. We suggest the following rationale for this result, examining how the probing and SCAO learn to predict hallucinations

SCAO and probing are fundamentally similar. Probing directly utilizes the raw hidden state of $\theta$, while SCAO focuses on the last hidden state of $\theta$, which is projected onto the vocab embedding space.

\begin{figure}
\includegraphics[width=0.9\linewidth]{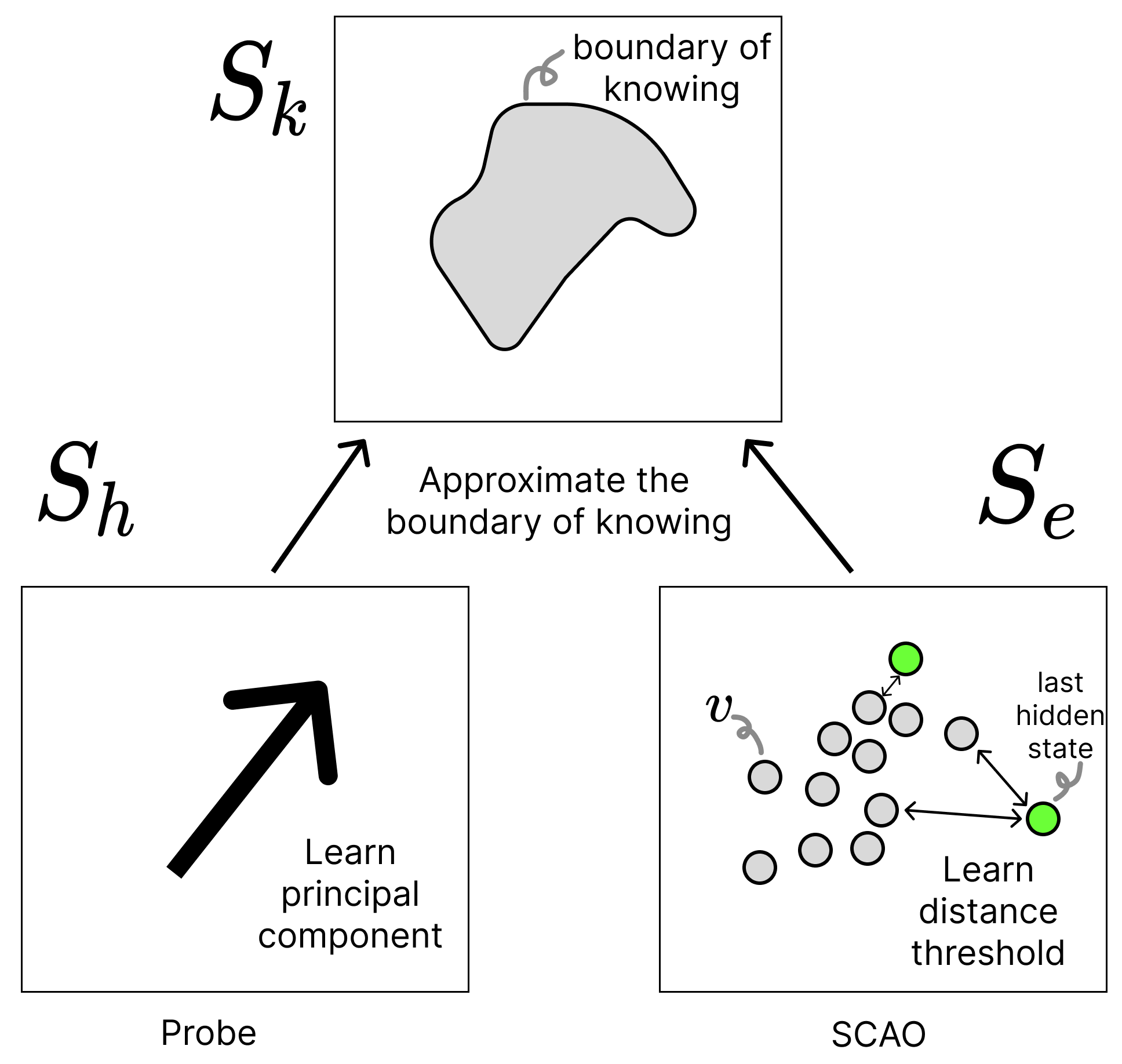}  
\caption{Illustration on two methods (probe, SCAO) approximating the boundary of knowing of $\theta$. In $S_e$ (lower right), the green balls are the last hidden state vector that is mapped to the vocab space. SCAO learns the threshold of distance between the hidden state and $v$ to classify $y$ of each ball.}
\label{fig:prob_scao}
\end{figure}

Let us assume a knowledge space (denoted as $S_k$) (\autoref{fig:prob_scao}), which represents the embedding of each knowledge in the $\theta$. And we term the gray area in the $S_k$ as a \textbf{boundary of knowing} of $\theta$, which represents the area where $k=True$ (model possesses the knowledge). This space is hypothetical and unknown but needs to be discovered to predict hallucination of $\theta$. To approximate this, what we have at hand are (1) the 4096-dimensional (in the case of \texttt{LLaMA-3-8B-Instruct}) hidden states (denoted as $S_h$) and (2) a vocab embedding space (denoted as $S_e$) of the same dimension, with vocab embedding vectors (denoted as $v$) distributed across $S_e$. 

In probing, a linear layer is trained to map $S_h$ to $S_k$. The weight of the linear layer is supposed to be a direction vector that represents a principal component of the boundary of knowing. Thus, an inner product with this vector tells if the given hidden states match the direction. Since it utilizes all 4096 dimensions to describe $S_k$, it offers high informational resolution, leading to generally strong performance.

Conversely, SCAO assumes that $S_e$ approximately aligns with $S_k$, which means the $v$ (gray balls in \autoref{fig:prob_scao}) aligns with the boundary of knowing (gray area in \autoref{fig:prob_scao}). SCAO figures the shape of $S_k$ by measuring the distance between the hidden state vector from the last layer (green balls in \autoref{fig:prob_scao}) and embedding vectors $v$. These mechanisms yield the following properties: (1) SCAO leverages $S_e$ and $S_h$, thus utilizing more information than probing, which only uses $S_h$. (2) However, this information is compressed into a single scalar value, distance, leading to lower information resolution, showing lower performance than the probe. 3) Despite the lower resolution, this simplification appears to enhance generalization. For instance, in out-of-domain scenarios, probing struggles with unfamiliar features in $S_h$, while SCAO effectively handles these novel features by employing its simplified distance-based measure.

Since probing and SCAO reflect slightly different aspects of $S_k$, combining these two methods in a feature fusion appears to provide an additional performance boost by leveraging their complementary strengths.

\section{Detail of datasets}
\label{app:dataset}

\subsection{Datasets and their refinement strategies} 

In this paragraph, we present details about the benchmark dataset for evaluating the hallucination prediction method: Mintaka \citep{sen2022mintakacomplexnaturalmultilingual}, ParaRel \citep{elazar2021pararel}, HaluEval \citep{li2023haluevallargescalehallucinationevaluation}, HotpotQA \citep{yang2018hotpotqadatasetdiverseexplainable}, and Explain. We also describe the refinement strategies applied to each dataset to reduce their AQE. ``+ type'' refers to refinements related to question types, while ``+ domain'' refers to refinements based on question domains, following \autoref{tab:main_exp}

\paragraph{Mintaka} 
Mintaka is a challenging multilingual QA dataset consisting of 20,000 question–answer pairs collected from MTurk contributors and annotated with corresponding Wikidata entities for both questions and answers.
Mintaka includes five types of question–answer pairs (entity, boolean, numerical, date, and string) and eight categories (movies, music, sports, books, geography, politics, video games, and history). Among multiple languages, we only use English question-answer pairs.

(1) + type : Among five types, we exclude boolean and numerical questions, following the discussion in \autoref{sec:case_study} and \autoref{app:case_study}.
(2) + domain : We randomly selected half of the domains (books, movies, music, sports) as the training and validation sets, and assigned the remaining domains to the test set.

\paragraph{ParaRel}
ParaRel is originally a dataset designed for masked language modeling, containing factual knowledge expressed through diverse prompt templates and relational types. We utilize the rearranged version by \citep{zhang2024rtuninginstructinglargelanguage}. 
 This version consists of 25,133 prompt-answer pairs across 31 domains. It is further divided into two parts: the first 15 domains are classified as in-domain data, and the remaining 16 domains are classified as out-of-domain. 
 
(1) + domain: In the original setting, in-domain data was used as the test set. In the refined setting, the test set consists of out-of-domain data.


\paragraph{HotpotQA} HotpotQA is a question-answering dataset where each instance consists of a question, label (types including entity, boolean, numerical), and reference documents. We utilize only the question and answer to fit the closed-book scenario. An example of the question is ``What government position was held by the woman who portrayed Corliss Archer in the film Kiss and Tell?'', paired with the label ``Chief of Protocol''. We use the development dataset as a test set, following \citep{zhang2024rtuninginstructinglargelanguage}.

(1) + type: We exclude the ``comparison'' type, as it consists of yes/no questions or questions that require choosing between two given options.

\paragraph{Explain}

We present a benchmark \textbf{Explain} to evaluate a model's ability to provide a descriptive answer to an open-ended question. Explain is an extended and refined version of an open-ended long-form dataset in the well-known and verified work of FActScore \citep{min2023factscorefinegrainedatomicevaluation}. In FActScore, a small dataset is devised to test the fact-checking pipeline for long-form QA. This dataset is created by appending prompts like “Tell me a bio of  <entity>”  to person names sourced from Wikipedia. However, its subjects are limited to only person names, and it includes only 500 entries. 

To address this, we developed Explain. Explain covers more general categories such as people, history, buildings, culture, etc (the entities from Mintaka), with the dataset size expanded to about 15000 entries. The prompt is "Please give me an explanation about <entity>", which follows the concept of the dataset in FActScore. All entities used as objects in the Explain setting are sourced from entity-type questions in the Mintaka dataset. 

(1) + domain: Since the entities in Explain are sourced from Mintaka, we adopt the same domain-splitting strategy as used in Mintaka.

\subsection{Data statistics}

We present data statistics of our main benchmarks, Mintaka and ParaRel. 
We also present examples of questions, categories, and statistics on Explain (\autoref{tab:explain_category}, \autoref{tab:data_stat})
 
\begin{table}[h!]

\centering
\caption{\#data in each benchmarks}
\label{tab:data_stat}
\resizebox{\linewidth}{!}{
\begin{tabular}{
 lllllll
}
\toprule
                   & ParaRel         & Mintaka & HaluEval & HotpotQA      & Explain\\ \hline 
Train          & 5575      & 7583   & 6000 & 8000  &  7583   \\
Valid          & 5584    & 1075    & 2000 &  2000     &  1075       \\ 
Test        & 13974     & 2152   & 2000 &   7405    &   2152 \\ \hline
\end{tabular}
}
\end{table}

\begin{table*}[h]
\centering
\caption{Examples of questions in Explain}
\label{tab:explain}
\resizebox{0.8\textwidth}{!}{
\begin{tabular}{
 lllllll
}
\toprule
Questions    & Entity  \\ \hline 

Please give me an explanation about ``A Game of Thrones''.     & A Game of Thrones  \\  
Please give me an explanation about ``Simone Biles''.         & Simone Biles    \\
Please give me an explanation about ``Winston Churchill''.     & Winston Churchill    \\ 
Please give me an explanation about ``Fyodor Dostoevsky''.     & Fyodor Dostoevsky    \\ 

Please give me an explanation about ``District 12''.     & District 12    \\ 

Please give me an explanation about ``The Battle of Gettysburg''.   & The Battle of Gettysburg    \\ \hline
\end{tabular}
}
\end{table*}

\begin{table}[h]

\centering
\caption{\#Data for the entity domains in Explain}
\label{tab:explain_category}
\resizebox{0.8\linewidth}{!}{
\begin{tabular}{
 lllllll
}
\toprule
        & Train & Dev & Test   \\ \hline 
Music & 914         &  139 & 273\\
History & 1059      &  149 & 296\\
Geography & 1033    &  144 & 306\\
Politics & 1036     &  143 & 300\\
Video games & 1057  &  150 & 302\\
Movies & 953        &  138 & 269\\
Books & 1020        &  140 & 283\\
Sports & 909        &  128 & 245 \\ \hline
\end{tabular}
}
\end{table}

\section{Experimental detail}
\label{app:additional_exp}

\subsection{Detail on the three main approaches}
\label{app:three_appoach}
In this section, we provide detailed descriptions of the three main approaches evaluated in \autoref{sec:experiment}.  
(1) confidence-based: We adopt a simplified method that takes the top-\( n \) softmax probabilities of the first answer token and applies a threshold. \( n \) and the threshold \( t \) are learnable $\phi$.

For the threshold-based discrimination, we first use the mean value of top-$n$ vocabulary ($s_j$ for $j_{th}$ token confidence in top-$n$ candidates) and apply the threshold, as depicted in \autoref{eq:threshold}. Here, the learnable parameter $\phi=\{t,n\}$ consists of a threshold ($t$) and the number of vocabulary candidates ($n$). During the training session, every possible pair of $n$ and threshold ($n$ is 1 to 30 in 30 steps, $t$ is 0 to 0.1 in 3000 steps, total 90K $\{t,n\}$ pairs) are measured on the training dataset, and the pair with the highest accuracy is applied to the test session. 
\begin{eqnarray}
\phi(s) = \begin{cases}
1, & \text{if } \frac{1}{n}\sum_{j=1}^{n} s_j \geq t \\
0, & \text{if } \frac{1}{n}\sum_{j=1}^{n} s_j < t 

\end{cases} 
\label{eq:threshold} \end{eqnarray}

\begin{figure}
\centering
\includegraphics[width=0.7\linewidth]{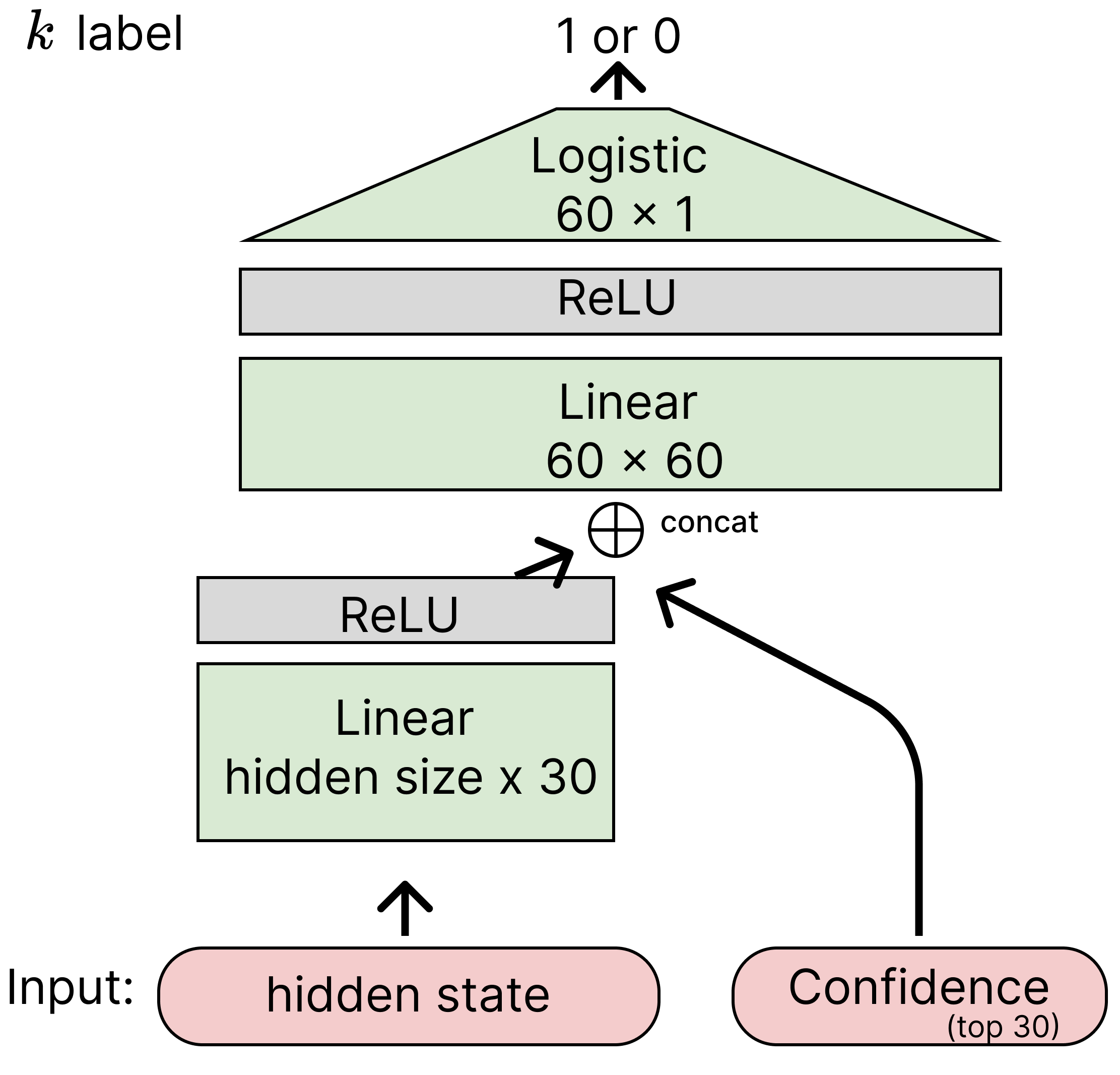}  
\caption{Structure of aggregation of hidden-state and confidence scores.}
\label{fig:fusion_structure}
\end{figure}

(2) hidden-state-based: We employ a 3-layer deep neural network (DNN) structure with dimensions \( d \rightarrow 100 \rightarrow 30 \rightarrow 1 \), where \( d \) is the hidden size. ReLU activation is applied between each layer. The objective function of DNN is binary cross entropy loss $L= -\frac{1}{N}\sum [y \cdot \log(\phi(s)) + (1-y)\cdot \log (1-\phi(s))]$. DNN ($\phi$) is trained on the dataset while $\theta$ is frozen. The choice of which layer's hidden state from $\theta$ to use is determined during the training phase, based on the one that achieves the highest validation accuracy.
This approach extends the linear probing \citep{li2024inference}.

We analyze that DNN emulates the mechanism of the mean threshold approach. The weights of the first layer decide how many candidates to count in, corresponding to the function of $n$ in the threshold-based approach. The second layer decides operations, such as mean or max pooling. DNN structure is a more suitable choice if feature fusion with other data is required.

(3) aggregation:  We utilize the feature fusion of confidence and hidden state. That implies utilizing both top-30 confidence value and  $h_{th}$ hidden state from $\theta$ as inputs to DNN.  This approach concatenates the confidence scores and hidden state into a single vector, which is then passed to a learnable \( \phi \) \citep{snyder2024early}. We use a module with dimensions of \( (d + n) \rightarrow 100 \rightarrow 30 \rightarrow 1 \), where $n$ is fixed to 30 (\autoref{fig:fusion_structure}).

\subsection{Experiment pipeline}
\label{app:exp_pipeline}
First the dataset is divided into $D_{train}$, $D_{valid}$, and $D_{test}$. We fit $\phi$ to $D_{train}$, while $\theta$ is frozen. The next step varies between the two types.

\paragraph{Learning-based}
The methods with hidden-state (\textit{Probe} and \textit{Probe + Conf}) should employ DNN architecture for $\phi$, which need machine learning. In this case, $\phi$ is trained on the $D_{train}$ with the objective of BCELoss. We train for 20 epochs to get $\phi$. And from the final checkpoint, we choose the index of the hidden layer of $\theta$ with the best accuracy on the $D_{valid}$. Then we use this hidden layer index and $\phi$ to test on the $D_{test}$. We calculate two metrics of accuracy and AUROC. When training, the learning rate is 1e-3, and the optimizer is AdamW.

\paragraph{Threshold-based}
The method that solely relies on confidence score (\textit{Conf}) uses a threshold for calibration. Here, learnable $\phi$ is the number of top candidate confidence scores ($n$) and the threshold ($t$). These two parameters are fitted in $D_{train}$, without evaluation on $D_{valid}$.
We select the $\phi$ that achieves the highest accuracy by performing a grid search over $t$ values in the range $[0,1]$ with 1000 uniformly spaced points, and $n$ in the range $[1,30]$ with 3000 uniformly spaced points. And use this $\phi$ to test on the $D_{test}$. AUROC is measured only with $n_{\phi}$, without $t_{\phi}$.

\subsection{AQE of larger model}

In this section, we provide the AQE results of the larger model (\texttt{LLaMA-3-70B-Instruct}) on the refined dataset, corresponding to \autoref{tab:aqe_refined}.

\begin{table*}[h!]
\caption{AQE score of dataset and \texttt{LLaMA3-70B-Instruct}. The version (original, type , domain) with the lowest AQE within each dataset is highlighted in \textbf{bold}.}
\label{tab:aqe_refined}
\resizebox{\textwidth}{!}{ \begin{tabular}{
  ccccccccccc
}
\toprule
\multicolumn{1}{l}{} & \multicolumn{3}{c}{Mintaka} & \multicolumn{2}{c}{HotpotQA} & \multicolumn{2}{c}{ParaRel} & \multicolumn{2}{c}{Explain} 
\\ \cmidrule(lr){2-4}  \cmidrule(lr){5-6}  \cmidrule(lr){7-8} \cmidrule(lr){9-10}
&      original    & + type      & + type + domain           & original    & + type  &  original & + domain &original & + domain   \\ \hline \addlinespace
$p(k=True)$   & 62.17 & 57.28 & 57.42 & 33.81 & 26.64 & 51.71 & 53.01 & 55.71 & 54.35 \\ 
$p(k=False)$  & 37.82 & 42.71 & 42.57 & 66.18 & 73.35 & 48.28 & 46.98 & 44.28 & 45.64 \\ 
AQE$_{acc}$   & 65.52  &  62.15 & \textbf{58.96} & \textbf{68.06} & 71.44 & 76.68 &\textbf{ 53.22} & 61.96 & \textbf{55.31 }\\
AQE$_{auc}$   & 65.75  &  64.35 & \textbf{58.77} & 63.78 & \textbf{54.56} & 85.98 & \textbf{53.29}  & 67.69 & \textbf{57.71}\\
\bottomrule
\end{tabular} }
\end{table*}

\subsection{Experiment with accuracy}
\label{app:exp_acc}
In this section, we present the performance and accuracy-based deltas of various hallucination prediction methods. The results exhibit trends similar to those observed when using AUROC as the evaluation metric.

    \begin{table*}[t!]
    \caption{Hallucination prediction performance (accuracy) of instruction-tuned 8B and 70B LLaMA models across multiple datasets.}
    \label{tab:main_exp}
    \subcaptionbox{Mintaka}{
    \resizebox{\textwidth}{!}{  \begin{tabular}{ccccccccccccc} 
    \toprule
    \multicolumn{1}{c}{} & \multicolumn{6}{c}{8B} & \multicolumn{6}{c}{70B} \\ 
    \cmidrule(lr){2-7} \cmidrule(lr){8-13}  
    \multicolumn{1}{c}{} 
      & \multicolumn{2}{c}{original} 
      & \multicolumn{2}{c}{+ type} 
      & \multicolumn{2}{c}{+ type + domain} 
      & \multicolumn{2}{c}{original} 
      & \multicolumn{2}{c}{+ type} 
      & \multicolumn{2}{c}{+ type + domain} \\ 
    \cmidrule(lr){2-3}  \cmidrule(lr){4-5}  \cmidrule(lr){6-7}  
    \cmidrule(lr){8-9}  \cmidrule(lr){10-11}  \cmidrule(lr){12-13}
    & auroc & $\mathcal{A}(\phi(s_M))$ 
    & auroc & $\mathcal{A}(\phi(s_M))$ 
    & auroc & $\mathcal{A}(\phi(s_M))$ 
    & auroc & $\mathcal{A}(\phi(s_M))$ 
    & auroc & $\mathcal{A}(\phi(s_M))$ 
    & auroc & $\mathcal{A}(\phi(s_M))$ \\ 
    \midrule \addlinespace
    Conf      &   62.75        &      -        &   60.71         &  -    &   61.54       &    -  & 68.39 & - & 65.53 & - & 64.65 & -   \\ 
    Conf (\textbf{SCAO}) &   67.35       &    -         &  65.78          &  -    &  \underline{66.75} & - & 70.53 & - &  68.15 & - & \underline{66.98} & -   \\ 
    Probe$_{dnn}$  &  70.55   &  7.05       &    68.87     & 9.06        &  66.68 & 7.64 & 73.32 & 7.8 & 69.73 & 7.58 & 66.12 & 7.16 \\ 
    Conf + Probe   &    \underline{71.38}   &  7.88    &  \underline{69.98}    & 10.17    &  65.12 & 6.08 & \textbf{74.21} & 8.69 & \textbf{71.44} & 9.29 & 65.99 & 7.03  \\
      Conf + Probe (\textbf{SCAO})       &    \textbf{71.96 }  &  8.46    &  \textbf{79.41}   & 10.68      &   \textbf{68.21} & 9.17  & \underline{74.14} & 8.62 & \underline{71.35} & 9.20 & \textbf{67.61} & 8.65 \\
    \bottomrule
    \end{tabular} } }
    \end{table*}

\begin{table}
\vspace{-10pt}
\centering
\subcaption{HotpotQA}

\resizebox{\linewidth}{!}{ \begin{tabular}{
  cccccccccccccccccccc
}
\toprule
\multicolumn{1}{l}{} & \multicolumn{4}{c}{8B}   & \multicolumn{4}{c}{70B} \\ \cmidrule(lr){2-5} \cmidrule(lr){6-9} 
\multicolumn{1}{l}{} & \multicolumn{2}{c}{original} & \multicolumn{2}{c}{+ type}   & \multicolumn{2}{c}{original} & \multicolumn{2}{c}{+ type}
\\ \cmidrule(lr){2-3}  \cmidrule(lr){4-5}  \cmidrule(lr){6-7}  \cmidrule(lr){8-9}  
               & acc        &$\mathcal{A}(\phi(s_M))$    & acc         & $\mathcal{A}(\phi(s_M))$   & acc        & $\mathcal{A}(\phi(s_M))$   & acc         & $\mathcal{A}(\phi(s_M))$       \\ \hline \addlinespace
Conf            &  71.54  &  -  &    75.93 & -  & 69.75 & - & 73.08 & - \\ 
 Conf (\textbf{SCAO})     &   73.23 &  -  &    \underline{76.93} & -  & 70.97 & - & \textbf{75.97} & -  \\
Probe$_{dnn}$   &   75.49   &  6.94    &    76.31     & 0.28  & 71.24 & 3.18 & 71.91 & 0.47 \\ 
Conf + Probe   &   \underline{76.00} &  7.45   &  75.46   & -0.56 & \underline{72.50} & 4.44 & 74.02 & 2.58 \\
   Conf + Probe (\textbf{SCAO}) &  \textbf{77.69}   &  9.14   & \textbf{76.95} & 0.92  & \textbf{72.51} & 4.45 & \underline{74.90} & 3.46 \\
\bottomrule
\end{tabular} } 
\subcaption{ParaRel}


\resizebox{\linewidth}{!}{ \begin{tabular}{
  cccccccccccccccccccc
}
\toprule
\multicolumn{1}{l}{} & \multicolumn{4}{c}{8B}   & \multicolumn{4}{c}{70B} \\ \cmidrule(lr){2-5} \cmidrule(lr){6-9} 
\multicolumn{1}{l}{} & \multicolumn{2}{c}{original} & \multicolumn{2}{c}{+ domain}   & \multicolumn{2}{c}{original} & \multicolumn{2}{c}{+ domain}
\\ \cmidrule(lr){2-3}  \cmidrule(lr){4-5}  \cmidrule(lr){6-7}  \cmidrule(lr){8-9}  
               & acc        &$\mathcal{A}(\phi(s_M))$    & acc         & $\mathcal{A}(\phi(s_M))$   & acc        & $\mathcal{A}(\phi(s_M))$   & acc         & $\mathcal{A}(\phi(s_M))$       \\ \hline \addlinespace
Conf        &   67.58      &    -  &   62.63   & - & 65.75 & - & 57.29 & -  \\ 
Conf (\textbf{SCAO})    &  66.67    &       -    &   67.88    &   -   & 66.83 & - & \textbf{69.19} & -   \\ 
Probe$_{dnn}$    & 80.52   &  7.26       &   66.82     & 11.73     & \textbf{82.29} & 5.61 & 65.24 & 12.02 \\ 
Conf + Probe    & \underline{80.64} &  7.38    & \underline{68.65}    & 13.56   & \underline{82.18} & 5.50 & 64.46 & 11.24   \\
Conf + Probe (\textbf{SCAO}) & \textbf{80.92}  &  7.66   &  \textbf{69.24}    & 14.15 & 81.84 & 5.16 & \underline{66.36} & 13.14 \\
\bottomrule
\end{tabular} } 
\subcaption{Explain}
\resizebox{\linewidth}{!}{ \begin{tabular}{
  ccccccccccccc
}
\toprule
\multicolumn{1}{l}{} & \multicolumn{4}{c}{8B}   & \multicolumn{4}{c}{70B} \\ \cmidrule(lr){2-5} \cmidrule(lr){6-9} 
\multicolumn{1}{l}{} & \multicolumn{2}{c}{original} & \multicolumn{2}{c}{+ domain}   & \multicolumn{2}{c}{original} & \multicolumn{2}{c}{+ domain}
\\ \cmidrule(lr){2-3}  \cmidrule(lr){4-5}  \cmidrule(lr){6-7}  \cmidrule(lr){8-9}  
               & acc        &$\mathcal{A}(\phi(s_M))$    & acc         & $\mathcal{A}(\phi(s_M))$   & acc        & $\mathcal{A}(\phi(s_M))$   & acc         & $\mathcal{A}(\phi(s_M))$       \\ \hline \addlinespace
Conf       &  57.66 & - & 60.29 & - & 55.71 & - & 53.31 & -  \\ 
Conf (\textbf{SCAO})    & 59.71 & - & 59.57 & - & 60.99 & - & 60.78 & -   \\ 
Probe$_{dnn}$   & 78.92 & 13.27 & \underline{63.31} & 2.1 & \underline{76.28} & 14.32 & 62.65 & 7.34 \\ 
Conf + Probe    & \underline{79.12} & 13.47 & 64.78 & 3.57 & 76.00 & 14.04 & \underline{63.10} & 7.79 \\
 Conf + Probe (\textbf{SCAO}) & \textbf{80.91} & 15.26 & \textbf{63.66} & 2.45 & \textbf{78.75} & 16.79 & \textbf{64.87} & 9.56 \\
\bottomrule
\end{tabular} } 
\end{table}


{\color{black} -}
\newpage
\section{Instruction prompts}
\label{app:instruction}
In this section, we compile the instructional prompts employed in our study. Terms marked with $\underline{\texttt{underline}}$ indicate placeholders that need to be filled with the corresponding content.
\newtcolorbox{boxA}{
    fontupper = \bf,
    boxrule = 1.5pt,
    colframe = black 
}

A. Normal instruction template
\begin{boxA}
[Question]:\{\underline{\texttt{question}}\} [Answer]:
\end{boxA}

B. SCAO instruction template
\begin{boxA}
[Question]: \{\underline{\texttt{question}}\} You must answer in only one word. [Answer]:
\end{boxA}

C. G-eval instruction template

\begin{boxA}
[Instruction] The text enclosed within triple backticks (```) is an agent's answer to a given question ([Question]). \par
Your task is to evaluate whether the agent's response is factually correct. \par
{\color{white} .}\par
1) Analyze and explain whether the answer contains any factual inaccuracies. \par
2) Then, classify the answer as either "True" (only factually correct contents) or "False" (containing any factually incorrect content). \par
{\color{white} .}\par
``` \par
[Question]:\{\underline{\texttt{question}}\} [Answer]:\{\underline{\texttt{answer}}\} \par
''' \par
\end{boxA}


\section{Usage of AI assistants}

In preparing this manuscript, we relied on AI-powered writing tools to refine sentence flow, fix grammatical mistakes, and improve readability. These assistants were used strictly for language polishing and played no role in shaping the technical content, research design, or experimental work. All scientific concepts, findings, and conclusions presented in this paper were fully developed and written by the researchers. The involvement of AI was limited to editorial support and did not influence the originality or intellectual contributions of the study.



\end{document}